\documentclass[smallcondensed]{svjour3}       

\expandafter\let\csname opt@amsmath.sty\endcsname\relax

\smartqed  
\usepackage{natbib}
\setcitestyle{aysep={}}
\usepackage{mathptmx}
\usepackage{textcomp}
\usepackage{xcolor}
\usepackage{amsmath,amssymb,amsfonts}
\usepackage{graphicx}
\usepackage[mathscr]{euscript}
\usepackage{algorithmicx}
\usepackage[ruled,linesnumbered,vlined,shortend]{algorithm2e}
\usepackage{graphicx}
\usepackage{booktabs}
\usepackage{subfigure}
\usepackage{multirow}
\usepackage{tabularx}
\usepackage{tablefootnote}
\usepackage{threeparttable}
\usepackage{caption}
\usepackage[hidelinks]{hyperref}
\usepackage{diagbox}
\usepackage{verbatim}
\usepackage{mdwlist}
\usepackage{makecell}

\hypersetup{%
    pdfborder = {0 0 0}
}
\def\BibTeX{{\rm B\kern-.05em{\sc i\kern-.025em b}\kern-.08em
    T\kern-.1667em\lower.7ex\hbox{E}\kern-.125emX}}

\newtheorem{defn}{Definition}

\newcommand{\fullmethod}[0]{Time-Aware Tensor Decomposition}

\newcommand{\method}[0]{\textsc{TATD}\xspace}
\newcommand{\methodz}[0]{\textsc{TATD-0}\xspace}

\newcommand{\alsa}[0]{ALS + Adam\xspace}

\newcommand{\cpals}[0]{\textsc{CP-ALS}\xspace}

\newcommand{\argmin}[1]{\underset{#1}{\operatorname{arg}\,\operatorname{min}}\;}
\newcommand{\T}[1]{\boldsymbol{\mathscr{#1}}}
\newcommand{\tensor}[1]{\boldsymbol{\mathscr{#1}}}
\newcommand{\mat}[1]{\mathbf{#1}}
\newcommand{\vect}[1]{\mathbf{#1}}

\newcommand{\A}[1]{\mat{A}^{(#1)}}

\SetKwInput{KwInput}{Input}
\SetKwInput{KwOutput}{Output}

\newcommand*{\QEDB}{\hfill\ensuremath{\square}}%

\newcommand{\bair}[0]{{Beijing Air Quality}\xspace}
\newcommand{\mair}[0]{{Madrid Air Quality}\xspace}
\newcommand{\indoor}[0]{{Indoor Condition}\xspace}
\newcommand{\radar}[0]{{Radar Traffic}\xspace}
\newcommand{\server}[0]{{Server Room}\xspace}

\newcommand{\hide}[1]{}

\begin{document}

\title{Time-Aware Tensor Decomposition for Missing Entry Prediction }

\author{Dawon Ahn 	\and 
		Jun-Gi Jang \and
		U Kang
}

\institute{Dawon Ahn \at
              Seoul National University, 
              Seoul, South Korea \\
              \email{dawon@snu.ac.kr} \and
           Jun-Gi Jang \at
              Seoul National University,
               Seoul, South Korea \\
              \email{elnino4@snu.ac.kr} \and
           U Kang \at
              Seoul National University, 
              Seoul, South Korea \\
              \email{ukang@snu.ac.kr}
}

\date{Received: date / Accepted: date}

\maketitle

\begin{abstract}
	Given a time-evolving tensor with missing entries, how can we effectively factorize it for precisely predicting the missing entries?
%
Tensor factorization has been extensively utilized for analyzing various multi-dimensional real-world data.
However, existing models for tensor factorization
have disregarded the temporal property for tensor factorization while most real-world data are closely related to time.
Moreover, they
do not address accuracy degradation due to the sparsity of time slices.
The essential problems of how to exploit the temporal property for tensor decomposition and consider the sparsity of time slices remain unresolved.

In this paper, we propose \method (\fullmethod), a novel tensor decomposition method for real-world temporal tensors. 
\method is designed to exploit temporal dependency and time-varying sparsity of real-world temporal tensors.
We propose a new smoothing regularization with Gaussian kernel for modeling time dependency.
Moreover, we improve the performance of \method by considering time-varying sparsity.
We design an alternating optimization scheme suitable for temporal tensor factorization with our smoothing regularization.
%
Extensive experiments show that \method provides the state-of-the-art accuracy for decomposing temporal tensors.

\end{abstract}

\section{Introduction}
\label{sec:introduction}

Given a temporal tensor where one mode denotes time,
how can we discover its latent factors for effectively predicting missing entries?
A tensor, or multi-dimensional array, has been widely used to model multi-faceted relationships for time-evolving data.
For example, air quality data~\Citep{zhang2017cautionary} containing measurements of contaminants collected from sensors at every time step are modeled as a 3-mode temporal tensor (time, site, contaminant).
%
Tensor factorization is a fundamental building block for effectively analyzing tensors by revealing latent factors between entities~\Citep{kolda2008scalable,KangPHF12,bahadori2014fast,OhPSK18,ParkOK17, liu2019costco},
and it has been extensively utilized in various real-world applications across diverse domains including recommender systems~\Citep{symeonidis2016matrix}, clustering~\Citep{sun2015heterogeneous}, anomaly detection~\Citep{kolda2008scalable}, correlation analysis~\Citep{sun2006window}, network forensic~\Citep{maruhashi2011multiaspectforensics}, and latent concept discovery~\Citep{KoldaBK05}.
CANDECOMP/PARAFAC (CP)~\Citep{harshman1970foundations} factorization is one of the most widely used tensor factorization models, which factorizes a tensor into a set of factor matrices and a core tensor which is restricted to be diagonal~\Citep{harshman1970foundations}.

Previous CP factorization methods~\Citep{kolda2008scalable,KangPHF12,bahadori2014fast,OhPSK18,ParkOK17,liu2019costco, symeonidis2016matrix, KoldaBK05, sun2006window, maruhashi2011multiaspectforensics, sun2015heterogeneous} do not consider temporal property when it comes to factorizing tensors while most time-evolving tensors exhibit temporal dependency.
Recently, few studies~\Citep{wu2019neural,yu2016temporal} have been conducted to address the above problem, but they consider only the past information of a time step to model temporal dependencies.
A model needs to examine both the past and the future information to capture accurate temporal properties since information at each time point is heavily related to that in the past and the future.
In addition, they do not consider the time-varying sparsity, one of the main properties in temporal tensors.	
The main challenges to design an accurate temporal tensor factorization method for missing entry prediction are
	1) how to harness the time dependency in real-world data,
 and
	2) how to exploit the varying sparsity of temporal slices.
%


In this paper, we propose \method (\fullmethod), a time-aware tensor factorization method for analyzing real-world temporal tensors.
	Our main observation is that adjacent time factor vectors are mostly similar to each other since time slices in a temporal tensor are closely related to each other. 
%
	Based on this observation, \method employs a kernel smoothing regularization to make time factor vectors reflect temporal dependency.
	Moreover, \method imposes a time-dependent sparsity penalty to strengthen the smoothing regularization.
	The sparsity penalty modulates the amount of the regularization using the sparsity of time slices.
	\method further improves accuracy using an effective alternating optimization scheme that incorporates an analytical solution and Adam optimizer.
	Through extensive experiments, we show that \method effectively considers the time dependency for tensor factorization, and achieves higher accuracy compared to existing methods.
	Our main contributions are as follows:
%
	\begin{itemize*}
	    \item {\textbf{Method.}
	    	We propose \method, a novel tensor factorization method considering temporal dependency.
	        \method exploits a smoothing regularization for effectively modeling time factor with time dependency, and adjusts it by utilizing a time-varying sparsity.
	    }
	    \item {\textbf{Optimization.}
	        We propose an alternating optimization strategy suitable for our smoothing regularization.
	        The strategy alternatively optimizes factor matrices with an analytical solution and Adam optimizer.
	    }
	    \item {\textbf{Performance.}
	        Extensive experiments show that exploiting temporal dependency is crucial for accurate tensor factorization of temporal tensors.
	        \method achieves up to $7.01\times$ lower RMSE and $5.50\times$ lower MAE for sparse tensor factorization compared to the second best methods. 
	        }
	\end{itemize*}

The rest of the paper is organized as follows.
In Section~\ref{sec:preliminaries}, we explain preliminaries on tensor factorization.
Section~\ref{sec:proposed} describes our proposed method~\method.
We demonstrate our experimental results in Section~\ref{sec:experiment}.
After reviewing related work in Section~\ref{sec:related},
we conclude in Section~\ref{sec:conclusion}.

\setlength{\extrarowheight}{0em}
\setlength{\tabcolsep}{1pt}
\begin{table}[t!]
    \centering
    \caption{Table of symbols}
   	\label{tab:symbols}
    \begin{tabular}{c  l}
		\toprule
		\textbf{Symbol} & \textbf{Definition} \\
		\midrule
        $\T{X}$ & input tensor $\in \mat{R}^{I_{1} \times ... \times I_{N}}$ \\
		$\alpha$ & index $(i_{1}, ..., i_{N})$ of $\T{X}$\\
        $x_{\alpha}$ & entry of $\T{X}$ with index $\alpha$ \\
        $N$ & order of tensor $\T{X}$ \\
        $I_{n}$ & length of the $n$th mode of tensor $\T{X}$\\
        $\mat{A}^{(n)}$ & $n$th factor matrix $(\in \mat{R}^{I_{n} \times K})$ \\
        $\vect{a}^{(n)}_{i_{n}}$ & $i_{n}$th row of $\mat{A}^{(n)}$ \\
        $a^{(n)}_{i_{n}k}$ & $(i_{n}, k)$th entry of $\mat{A}^{(n)}$ \\
        $K$ & rank of tensor $\T{X}$ \\
        $t$ & time mode of $\T{X}$  \\
        $\T{X}_{i_t}$ & $i_t$th time slice of size $I_1\times \cdots \times I_{t-1} \times I_{t+1}\cdots \times I_{N}$\\
        $\omega_{i_t}$ & number of observed entries of time slice $\T{X}_{i_t}$ \\
        $\|\tensor{X}\|_F$ & Frobenius norm of tensor $\T{X}$\\
		$\lambda_t$, $\lambda_r$ & regularization parameter \\
        \bottomrule
    \end{tabular}
\end{table}

\section{Preliminaries}
\label{sec:preliminaries}

We describe the preliminaries of tensor and tensor decomposition.
We use the symbols listed in Table~\ref{tab:symbols}.

\subsection{Tensor and Notations} \label{sec:prelim:tensor}
Tensors are defined as multi-dimensional arrays that
generalize the one-dimensional arrays (or vectors) and
two-dimensional arrays (or matrices) to higher dimensions.
Specifically, the dimension of a tensor is referred to as order or way;
the length of each mode is called ‘dimensionality’ and denoted by $I_1, \cdots, I_n$.
%
We use boldface Euler script letters (e.g., $\T{X}$) to denote tensors,
boldface capitals (e.g., $\mat{A}$) to denote matrices,
and
boldface lower cases (e.g., $\vect{a}$) to denote vectors.
%
The $\alpha = (i_1, \cdots, i_N)$th entry of tensor $\T{X}$ is denoted by $x_{\alpha}$.

A slice of a 3-order tensor is a two-dimensional subset of it.
There are the horizontal, lateral, and frontal slices in a 3-order tensor $\tensor{X}$, denoted by $\mat{X}_{i_1::}$, $\mat{X}_{:i_2:}$, and $\mat{X}_{::i_3}$.
A tensor containing a mode representing time is called a \textit{temporal tensor}.

A time slice in a $3$-mode temporal tensor represents a two-dimensional subset disjointed by each time index.
For example, $\mat{X}_{i_t::}$ is an $i_t$th time slice when the first mode is the time mode.
For brevity, we express $\mat{X}_{i_t::}$ as $\mat{X}_{i_t}$.
Our proposed method is not limited to a $3$-mode tensor so that a time slice in an N-order temporal tensor corresponds to an (N-1)-dimensional subset of the tensor sliced by each time index.
We formally define a time slice $\T{X}_{i_{t}}$ as follows:

\begin{defn}[Time slice $\T{X}_{i_{t}}$]
\label{def:timeslice}	
	{Given an $N$-order tensor $\T{X} \in \mathbb{R}^{I_1\times \cdots \times I_{N}}$ and a time mode $t$,
	 we extract time slices of size $I_1\times \cdots \times I_{t-1} \times I_{t+1}\cdots \times I_{N}$ by slicing the tensor $\T{X}$ 
	so that an $i_{t}$th time slice $\T{X}_{i_{t}} \in \mathbb{R}^{I_1\times \cdots \times I_{t-1} \times I_{t+1}\cdots \times I_{N}}$ is an $N-1$ order tensor obtained at time $i_{t}$  where $1 \leq i_{t} \leq I_{t}$. \QEDB}
\end{defn}
%

%
The \textit{Frobenius norm} of a tensor $\T{X}$ $(\in \mat{R}^{I_{1} \times ... \times I_{N}})$ is given by
$ ||\T{X}||_F = \sqrt{\sum_{\alpha \in \Omega}{\T{X}^{2}_{\alpha}}}$, where $\Omega$ is the set of indices of entries in $\T{X}$, $ \alpha\ = (i_1,\cdots, i_N)$ is an index included in $\Omega$,
and $\T{X}_{\alpha}$ is the $(i_1, \cdots, i_N)$th entry of the tensor $\T{X}$.

\subsection{Tensor Decomposition} \label{sec:prelim:cp}
\begin{figure*}
    \centering
   	\vspace{5mm}
      \includegraphics[width=0.5\linewidth]{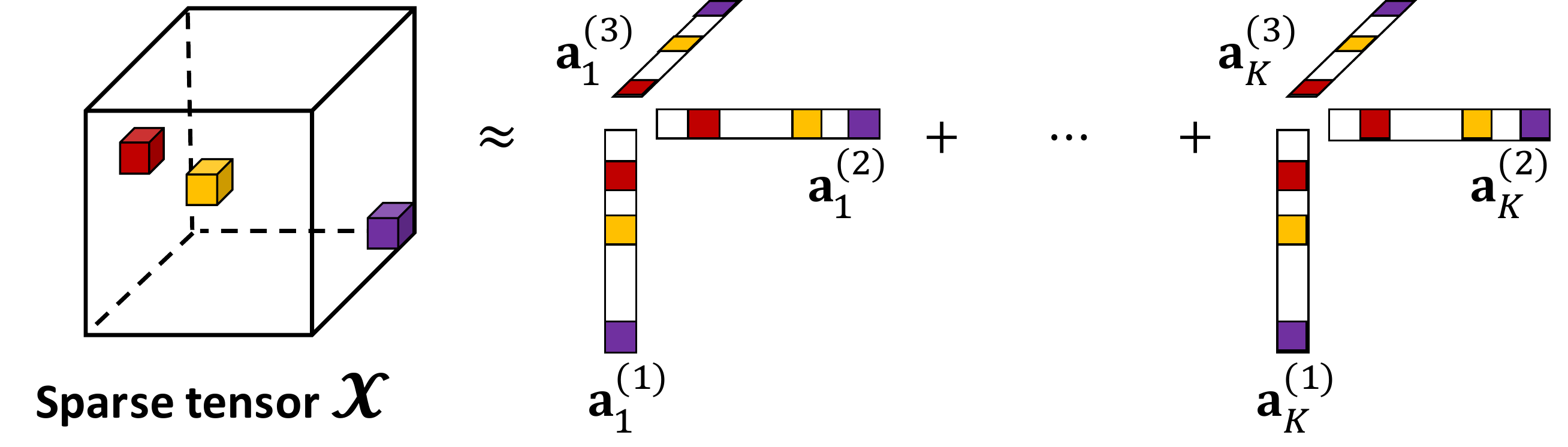}
    \caption{CP decomposition of a 3-way sparse tensor into $K$ components}
    \label{fig:3way_cp}
\end{figure*}

We provide the definition of CP 
decomposition~\Citep{harshman1970foundations,kiers2000towards}
which is one of the most representative factorization models.
Fig.~\ref{fig:3way_cp} illustrates CP decomposition of a $3$-way sparse tensor.
Our model \method is based on CP decomposition.
%
\begin{defn}[CP decomposition] \label{def:tensor_decomp}
	Given a 
	rank $K$ and an $N$-mode tensor $\T{X} \in \mathbb{R}^{I_{1}
	    \times \cdots \times I_N}$ with observed entries,
	CP decomposition approximates $\T{X}$ by finding latent factor matrices $\{\A{n} \in \mathbb{R}^{I_n \times K}\:|\:1 \leq n \leq N\}$.
	The factor matrices are obtained by minimizing the following loss function:
	\begin{align}
    	\T{L} \left(\A{1}, \cdots, \A{N}\right)
    		& = \sum_{\forall \alpha \in \Omega}{ \left({x}_{\alpha}-\sum_{k=1}^{K} \prod_{n=1}^{N}a^{(n)}_{i_{n}k}\right)^{2}} \label{eq:tensor_decomp:obs}
    \end{align}
	 where $\Omega$ indicates the set of the indices of the observed entries, $x_\alpha$ indicates the $\alpha = (i_1, \cdots, i_N)$th entry of $\T{X}$,
	 and $a^{(n)}_{i_{n}k}$ indicates $(i_{n}, k)$th entry of $\A{n}$.  \QEDB
\end{defn}


The standard CP decomposition method is not specifically designed to deal with temporal dependency;
thus CP decomposition does not give an enough accuracy for predicting missing values in a temporal tensor.
Although a few methods~\Citep{yu2016temporal,wu2019neural} have tried to capture temporal interaction,
none of them 1) captures temporal dependency between adjacent time steps, and 2) exploits the sparsity of temporal slices.
Our proposed \method carefully captures temporal information and considers sparsity of temporal slices for better accuracy in decomposing temporal tensors.

\section{Proposed Method}
\label{sec:proposed}

\begin{figure*}
	\vspace{2mm} 
	\centering{
		\includegraphics[width=\linewidth]{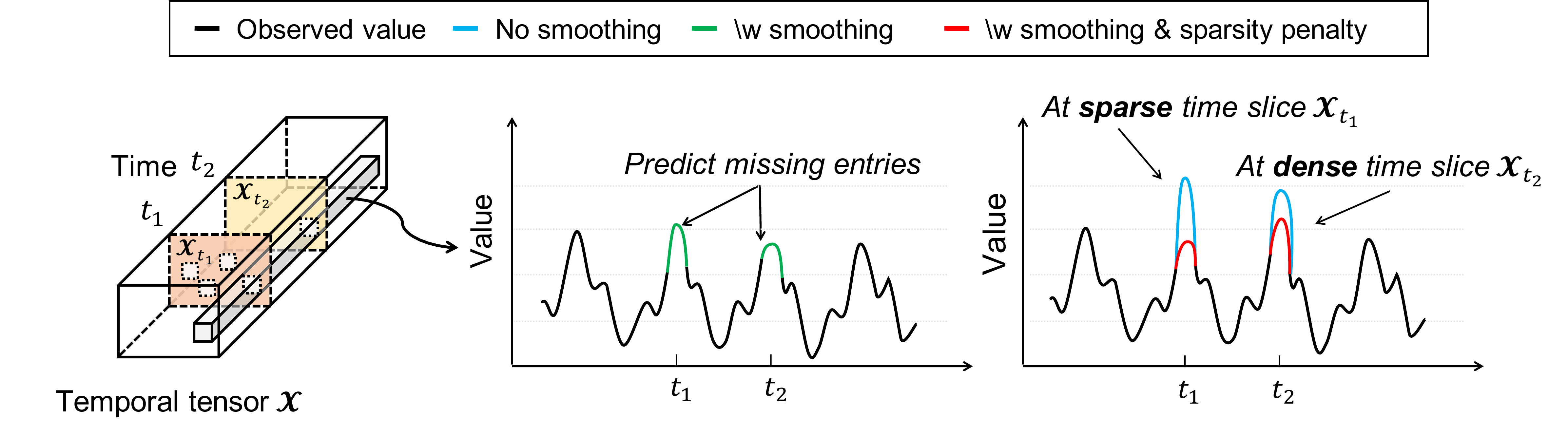}
	}
	\caption{\label{fig:overview}
		Illustration of a smoothing regularization and sparsity penalty by \method.
	 	\method accurately predicts missing entries of time slices 
	 		at $t_1$ and $t_2$ with the smoothing regularization.
	    Also, \method adjusts the amount of the smoothing regularization 
	    	according to the time-varying sparsity.
	    For a sparse time slice at $t_1$,
	    	\method predicts its missing entries via strong regularization 
	    	to actively consider nearby slices.
	   	For a dense time slice at $t_2$, 
	   		\method predicts its missing entries with weak regularization 
	   		by paying little attention to nearby slices
	}
\end{figure*}

In this section, we propose \method (Time-Aware Tensor Decomposition), 
	a tensor factorization method for temporal tensors.
We first introduce the overview of \method in Section~\ref{sec:overview}.
We then explain the details of \method in Sections~\ref{sec:sm} and~\ref{sec:sp},
and the optimization technique in Section~\ref{sec:opt}.

\subsection{Overview} \label{sec:overview}
	\method is a tensor factorization method designed for temporal tensors with missing entries.
	There are several challenges in designing an accurate tensor factorization method for temporal tensors.

	\begin{enumerate}
		\item {\bf Model temporal dependency.}
			Temporal dependency is an essential structure of temporal tensor.
			How can we design a tensor factorization model to reflect the temporal dependency?
		\item {\bf Exploit sparsity of time slices.}
			Time-evolving tensor has varying sparsity for its temporal slices.
			How can we exploit the temporal sparsity for better accuracy?
		\item {\bf Optimization.}
            How can we efficiently train our model and minimize its loss function?
	\end{enumerate}
	To overcome the aforementioned challenges, we propose the following main ideas.

	\begin{enumerate}
		\item {\bf Smoothing regularization (Section~\ref{sec:sm}).}
			We propose a smoothing regularization on time factor to capture temporal dependency.
		\item {\bf Time-dependent sparsity penalty (Section~\ref{sec:sp}).}
			We propose a time-dependent sparsity penalty to further improve the accuracy.
		\item {\bf Careful optimization (Section~\ref{sec:opt}).}
			We propose an optimization strategy utilizing an analytical solution and Adam optimizer to efficiently and accurately train our model.
	\end{enumerate}
Fig.~\ref{fig:overview} illustrates overview of \method.
We observe that adjacent time slices in a temporal tensor are closely related with each other due to temporal trend of the tensor.
Based on the observation, \method uses smoothing regularization such that time factor vectors for adjacent time slices become similar.
We also observe that different time slices have different densities.
Instead of applying the same amount of regularization for all the time slices,
we control the amount of regularization based on the sparsity of time slices
such that sparse slices are affected more from the regularization.
It is also crucial to efficiently optimize our objective function.
We propose an optimization strategy exploiting alternating minimization to expedite training and improve the accuracy.

\subsection{Smoothing Regularization}\label{sec:sm}
We describe how we formulate the smoothing regularization on tensor decomposition to capture temporal dependency.
Our main observation is that temporal tensors have temporal trends, and
adjacent time slices are closely related.
For example, consider an air quality tensor containing measurements of pollutant at a specific time and location;
it is modeled as a 3-mode tensor $\T{X}$ (time, location, type of pollutants; measurements).
Since the amount of pollutants at nearby time steps are closely related,
the time slice $\T{X}_{t}$ at time $t$ is closely related to the time slices $\T{X}_{t-1}$ at time $t-1$ and $\T{X}_{t+1}$ at time $t+1$.
This implies the time factor matrix after tensor decomposition should have related rows for adjacent time steps.

Based on the observation, our objective function is as follows.
Given an $N$-order temporal tensor $\T{X} \in \mathbb{R}^{I_1 \times \cdots \times I_N}$ with observed entries $\Omega$,
the time mode $t$,
and a window size $S$,
we find factor matrices {$\mat{A}^{(n)} \in \mathbb R^{I_{n}\times K}$} {$,1 \leq n \leq N $} that minimizes
		\begin{align} \label{eq:method:loss}
				\T{L}  = \sum_{\alpha \in \Omega}{\left({x}_{\alpha}-\sum_{k=1}^{K} \prod_{n=1}^{N}a^{(n)}_{i_{n}k}\right)^{2}}
				   + \lambda_t \sum_{i_{t}=1}^{I_{t}}\lVert{\vect{a}_{i_{t}}^{(t)}}-{{\tilde{\vect{a}}_{i_{t}}^{(t)}}}\rVert_{\text{2}}^{2} 
				   + \lambda_r\sum_{n \neq t}^{N}\lVert{\mat{A}^{(n)}}\rVert_{\text{F}}^{2}
		\end{align}
	where we define
		\begin{equation} \label{eq:method:smooth}
			{\tilde{\vect{a}}_{i_t}^{(t)}} = \sum_{i_s \in \mathscr{N}({i_t}, S)}{w(i_t,i_s)}{\vect{a}_{i_s}^{(t)}},
		\end{equation}
	and $\mathscr{N}(i_t, S)$ indicates adjacent indices $i_s$ of $i_t$ in a window of size $S$.
$\lambda_t$ and $\lambda_r$ are regularization constants 
	to adjust the effect of time smoothing and weight decay, respectively.
$\tilde{\vect{a}}_{i_t}^{(t)}$ in Equation~\eqref{eq:method:smooth} 
	denotes the smoothed row of the temporal factor.
The $\sum_{i_{t}=1}^{I_{t}}\lVert{\vect{a}_{i_{t}}^{(t)}}-{{\tilde{\vect{a}}_{i_{t}}^{(t)}}}\rVert_{\text{2}}^{2}$ term in Equation~\eqref{eq:method:loss} means 
	that we regularize the $i_t$th row of the temporal factor 
	to the smoothed vector from the neighboring rows in the factor.
The weight $w(i_t, i_s)$ denotes the weight to give to the $i_s$th row of the temporal factor matrix for the smoothing the $i_t$th row of the temporal factor.

An important question is, how to determine the weight $w(i_t, i_s)$?
We use the Gaussian kernel for the weight function due to the following two reasons.
First, it does not require any parameters to tune, 
	and thus we can focus more on learning the factors in tensor decomposition.
Second, it fits our intuition that a row closer to the $i_t$th row should be given a higher weight.
In Section~\ref{sec:experiment}, we show that \method with Gaussian kernel outperforms all the competitors; however we note that other weight function can possibly replace the Gaussian kernel to further improve the accuracy, and we leave it as a future work.

Given a target row index ${i_t}$, an adjacent row index ${i_{s}}$, and a window size $S$,
the weight function based on the Gaussian kernel is as follows:
	\begin{equation} \label{eq:method:kernel}
		w(i_{t},i_{s}) = \frac{\mathscr{K}(i_{t},i_{s})}{{\sum_{i_{s'} \in \mathscr{N}({i_t}, S)}} \mathscr{K}(i_{t}, i_{s'})}
	\end{equation}
	where $\mathscr{K}$ is defined by 
		$$ \mathscr{K}(i_{t}, i_{s}) =\exp\left(-\frac{(i_{t}-i_{s})^2}{2\sigma^2}\right)$$
Note that $\sigma$ affects the degree of smoothing; a higher value of $\sigma$ imposes more smoothing.
For each $i_t$th time slice, 
	the model constructs a smoothed time factor vector  $\tilde{\vect{a}}_{i_t}$ based on nearby factor vectors $\vect{a}_{i_s}$ and the weights $w(i_{t},i_{s})$. 
Our model then aims to reduce the smoothing loss between the time factor vector $\vect{a}_{i_{t}}^{(t)}$ and the smoothed one ${{\tilde{\vect{a}}_{i_{t}}^{(t)}}}$. 

\subsection{Sparsity Penalty} \label{sec:sp}

\begin{figure*}
\centering{
	\subfigure[ \bair]{
		\includegraphics[width=0.3\linewidth]{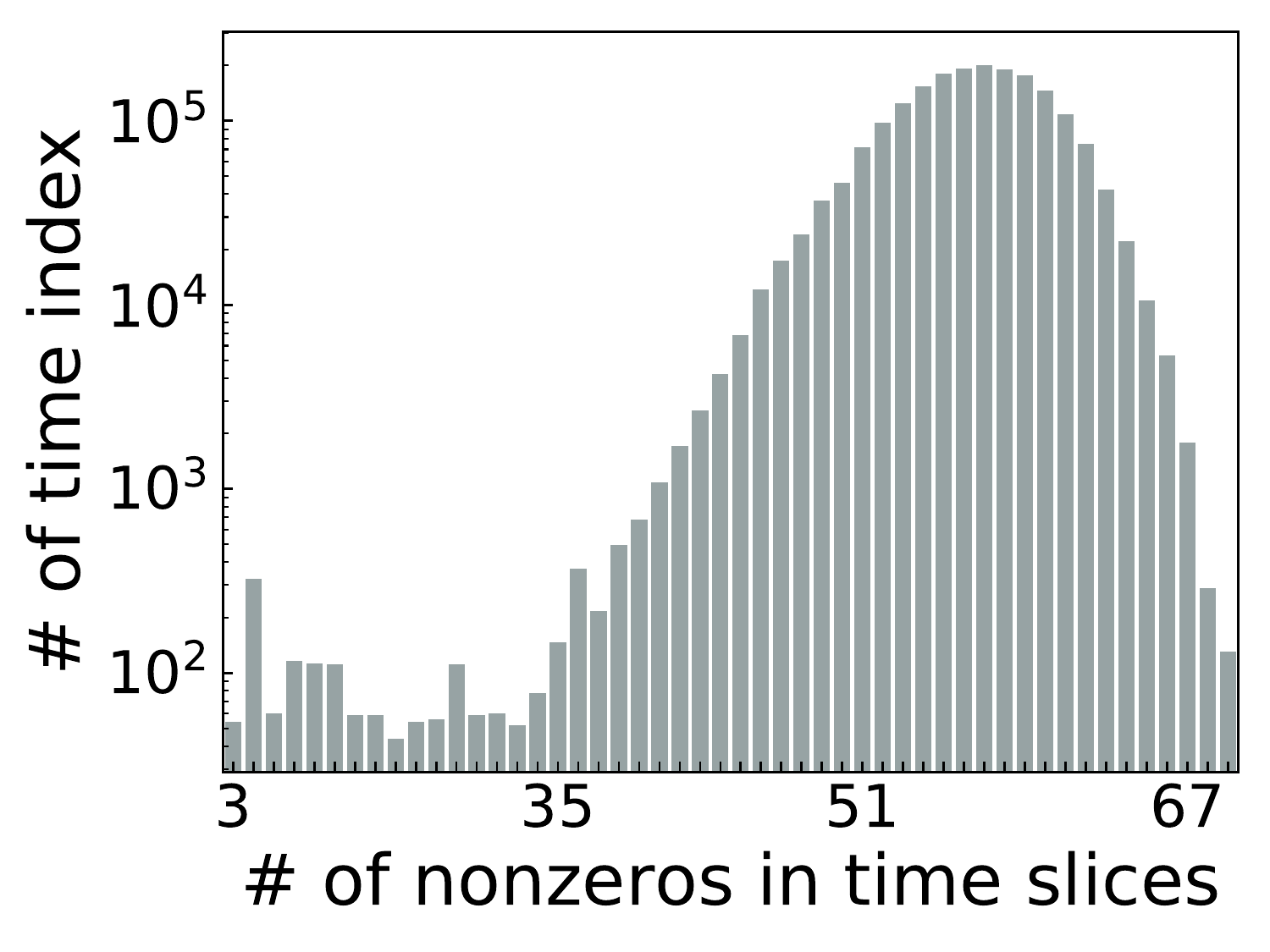}
	}
	\hspace{-3mm}
	\subfigure[ \mair]{
		\includegraphics[width=0.3\linewidth]{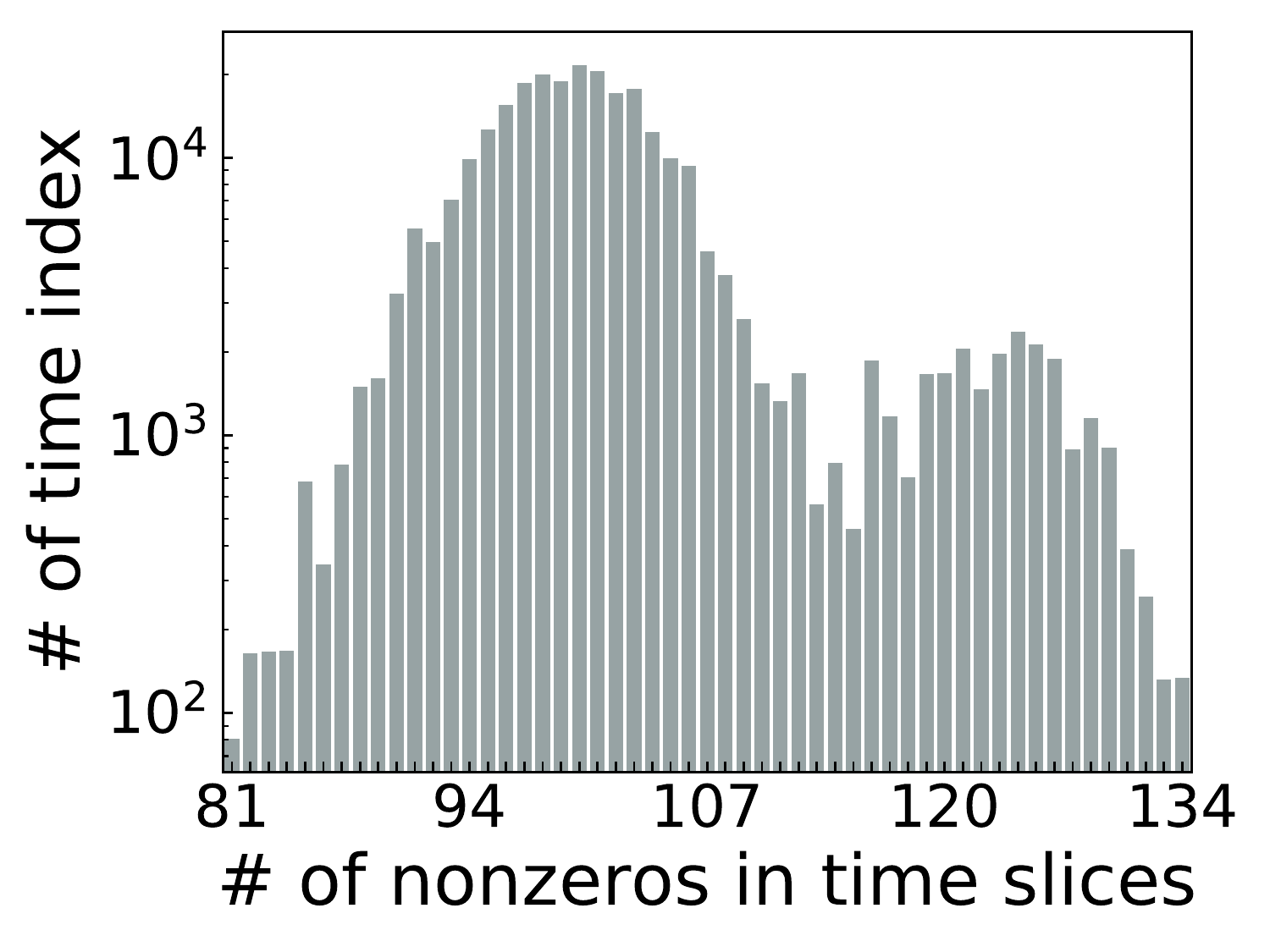}
	}
	\hspace{-3mm}
	\subfigure[ \radar]{
		\includegraphics[width=0.3\linewidth]{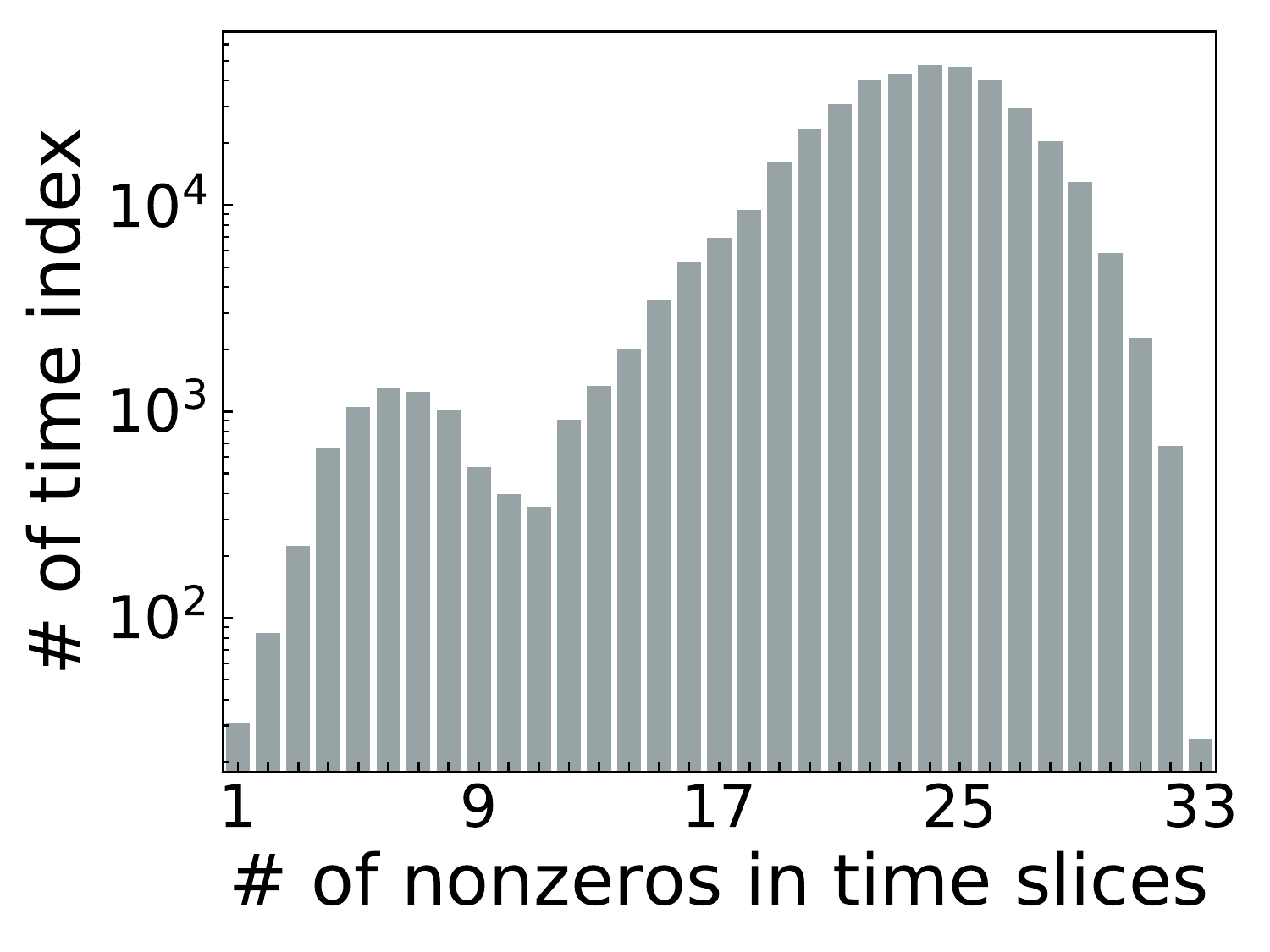}
	}\\
	\hspace{-3mm}
	\subfigure[ \indoor]{
		\includegraphics[width=0.3\linewidth]{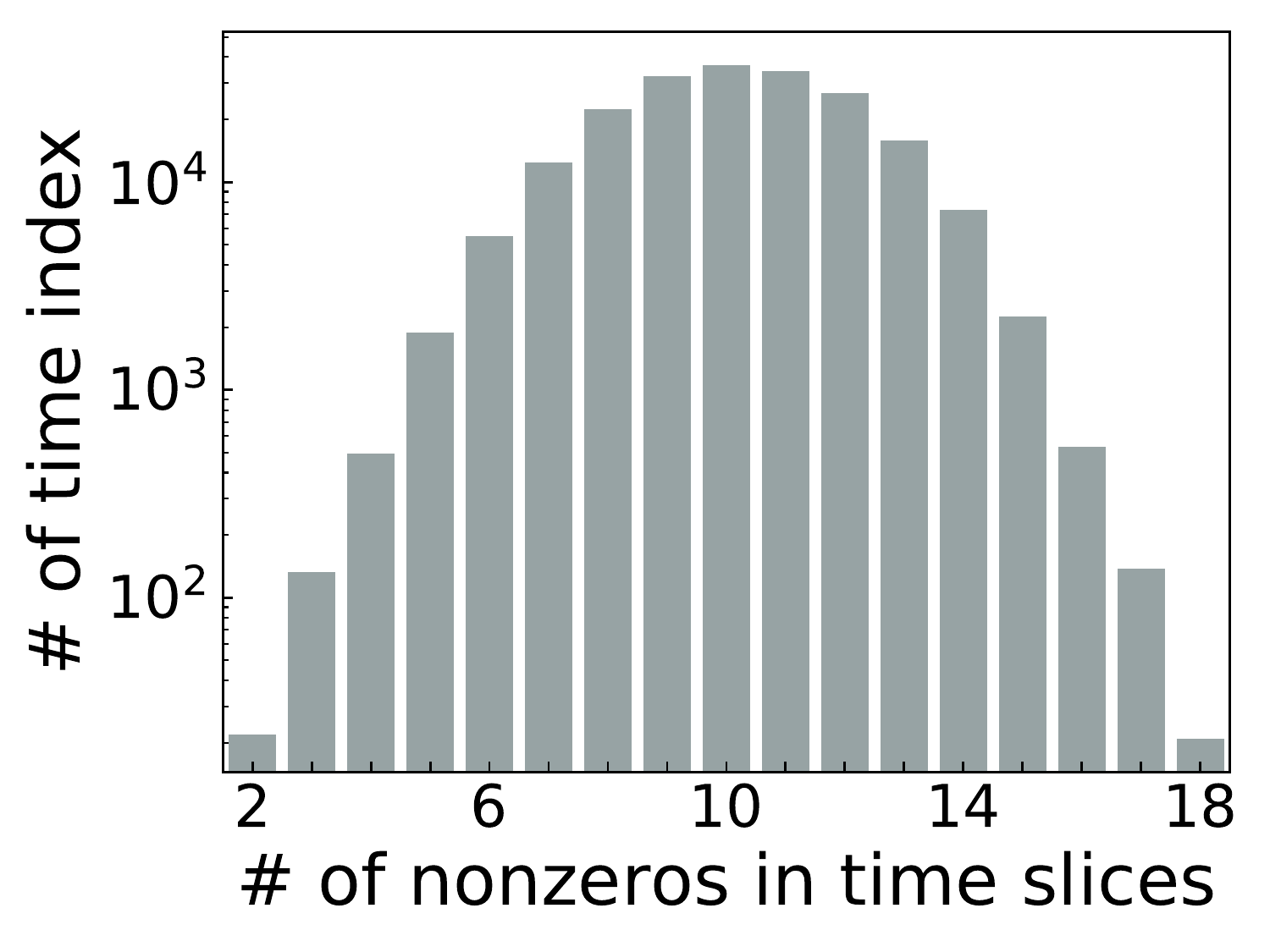}
	}
	\hspace{-3mm}
	\subfigure[ \server]{
		\includegraphics[width=0.3\linewidth]{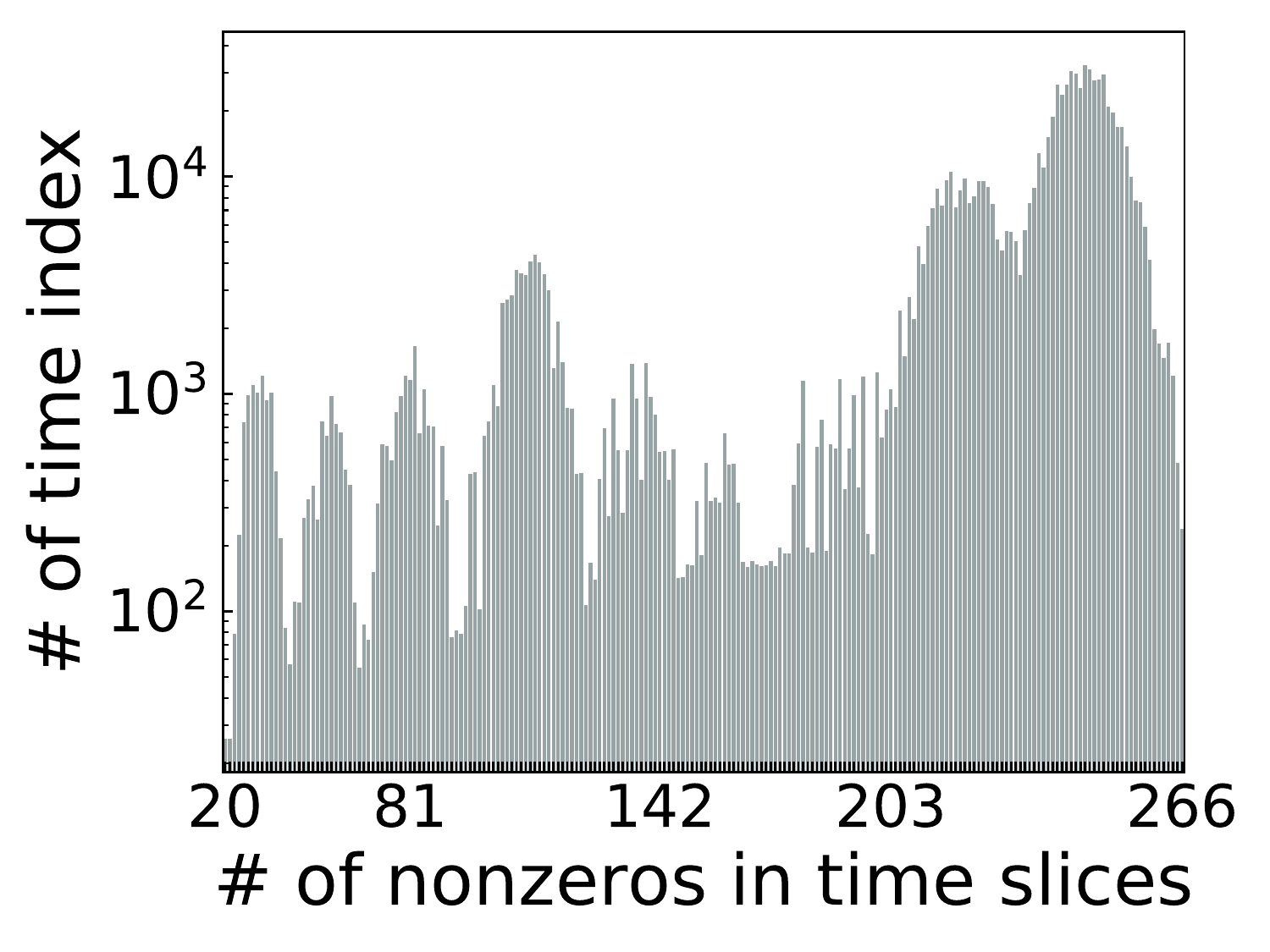}
	}
}
\caption{\label{fig:density}
	Time-varying density of five real-world datasets.
	The horizontal axis represents the unique number of nonzero entries in time slices.
	The vertical axis represents the number of time indices with such number of nonzeros.
	Note that time slices have varying densities
}
\end{figure*}

We describe how to further improve the accuracy of our method by considering the sparsity of time slices.
The loss function~\eqref{eq:method:loss} uses the same smoothing regularization penalty $\lambda_t$ for all the time factor vectors.
However, different time slices have different sparsity due to the different number of nonzeros in time slices (see Fig.~\ref{fig:density}), 
	and it is thus desired to design our method so that it controls the degree of regularization penalty depending on the sparsity.
For example, consider the 3-mode air quality tensor $\T{X}$ (time, location, type of pollutants; measurements), introduced in Section~\ref{sec:sm}, containing measurements of pollutant at a specific time and location.
Assume that the time slice $\T{X}_{t_1}$ at time $t_1$ is very sparse containing few nonzeros,
while the time slice $\T{X}_{t_2}$ at time $t_2$ is dense with many nonzeros.
The factor row $\vect{a}_{i_{t_2 }}^{(1)}$ at time $t_2$ can be updated easily using its many nonzeros.
However, the factor row $\vect{a}_{i_{t_1}}^{(1)}$ at time $t_1$ does not have enough nonzeros at its corresponding time slice, and thus it is hard to train $\vect{a}_{i_{t_1}}^{(1)}$
using only its few nonzeros;
we need to actively use nearby slices to make up for the lack of data.
Thus, it is desired to impose more smoothing regularization at time $t_1$ than at time $t_2$.

Based on the motivation, \method controls the degree of smoothing regularization based on the sparsity of time slices.
Let the \textit{time sparsity} $\beta_{i_{t}}$ of the $i_t$th time slice be defined as
	\begin{align} \label{eq:method:time_sparse}
		\beta_{i_{t}} &  =  1 - d_{i_t}
	\end{align}	
	where a \textit{time density} $d_{i_t}$ is defined as follows:
		\begin{align} \label{eq:method:time_norm_dense}
			d_{i_t} & = (0.999-0.001)  \frac{ \omega_{i_{t}} - \omega_\textit{min}}{\omega_\textit{max} - \omega_\textit{min}} + 0.001
	   \end{align}	
$\omega_{i_{t}}$ indicates the number of nonzeros at $i_{t}$th time slice;
$\omega_\textit{max}$ and $\omega_\textit{min}$ are the maximum and the minimum values of the number of nonzeros in time slices, respectively.
The \textit{time density} $d_{i_t}$ can be thought of as a min-max normalized version of $\omega_{i_{t}}$,
with its range regularized to [$0.001$, $0.999$].

Using the defined time sparsity, we modify our objective function as follows.
	\begin{align} \label{eq:method:naive_sp}
 			\T{L}  = \sum_{\alpha \in \Omega}{\left({x}_{\alpha}-\sum_{k=1}^{K} \prod_{n=1}		^{N}a^{(n)}_{i_{n}k}\right)^{2}}
		  		  +\sum_{i_{t}=1}^{I_{t}}\lambda_t \beta_{i_{t}} ||{{\vect a}_{i_{t}}^{(t)}}-{{\tilde{\vect{a}}_{i_{t}}^{(t)}}}||_{2}^{2} 
				  + \lambda_r\sum_{n \neq t}^{N}\lVert{\mat{A}^{(n)}}\rVert_{\text{F}}^{2}
	\end{align}

Note that the second term is changed to include the time sparsity $\beta_{i_{t}}$;
this makes the degree of the regularization vary depending on the sparsity of time slices.

Given the modified objective function in Equation~\eqref{eq:method:naive_sp}, we focus on minimizing the difference between $\vect{a}_{i_t}^{(t)}$ and $\tilde{\vect{a}}_{i_t}^{(t)}$ for time slices with a high sparsity rather than those with a low sparsity.
\method actively exploits the neighboring time slices when a target time slice is sparse, while it less exploits the neighboring ones for a dense time slice.


\subsection {Optimization} \label{sec:opt}

 \begin{algorithm}[t]
 	\caption{Training \method} \label{alg:als_grad}
 	\SetKwInOut{Input}{Input}
 	\SetKwInOut{Output}{Output}
 	\Input{
 		Tensor $\T{X} \in \mat{R}^{I_1 \times I_2 \times \cdots \times I_N}$ with observed entries $\forall \alpha = (i_1, \dots, i_n) \in \Omega $, rank $K$,
 		     window size $S$, sparsity penalty $\lambda_t$$\beta_{i_t}$,
 		     regularization parameter $\lambda_r$, and learning rate $\eta$\\
 	}
 	\Output{
 		Updated factor matrices $\mathbf{A}^{(n)} \in {R}^{I_n \times k}$ $ (n = 1...N)$
 	}
 	\vspace{1.5mm}
 	initialize all factor matrices $\mathbf{A}^{(n)}$ for $n = 1...N$ \label{alg:proposed:init}
 	
 	\Repeat{convergence criterion is met}
 	{
			\For{$n=1...N$}
			{
				loss $\T{L}$ $\leftarrow$  Eq.~\eqref{eq:method:naive_sp}  \label{alg:proposed:inner_start} \\
 			\If{$n$ is a time mode}{
				\Repeat{convergence criterion is met}
					{update a factor $\mathbf{A}^{(n)}$ using Adam optimizer with a learning rate $\eta$ \label{alg:proposed:timefactor}}
     		}
     		\Else{
				\For{$i_n = 1...I_n$}{
	     		update a row factor $\vect{a}_{i_n}^{(n)}$ using the row-wise update rule
	     		  \label{alg:proposed:inner_end}
	     		}		
     		}
		}
 	}
 \end{algorithm}

To minimize the objective function in Equation~\eqref{eq:method:naive_sp},
\method uses an alternating optimization method;
it updates one factor matrix at a time while fixing all other factor matrices.
%
\method updates non-time factor matrices using the row-wise update rule~\Citep{shin2016fully} while updating the time factor matrix using the Adam optimizer~\Citep{kingma2014adam}.
It allows \method to quickly converge with a low error, compared to naive gradient-based methods.



%

{\bf{Updating non-time factor matrix.}}
We note that updating a non-time factor matrix while fixing all other factor matrices is solved via the least square method, and we use the row-wise update rule~\Citep{shin2016fully,OhPSK18} in ALS for it.
The row-wise update rule is advantageous since it 
gives the optimal closed-form solution, and allows parallel update of factors.
We describe the details of the row-wise update rule in Appendix~\ref{subsec:rowwiseupdate}.


{\bf{Updating time factor matrix.}}
Updating the time factor matrix while fixing all other factor matrices is not the least square problem any more, and thus we turn to gradient based methods.
We use the Adam optimizer which has shown superior performance for recent machine learning tasks.
We verify that using the Adam optimizer only for the time factor leads to faster convergence compared to other optimization methods in Section~\ref{sec:experiment}.

{\bf{Overall training.}}
Algorithm~\ref{alg:als_grad} describes how we train \method.
We first initialize all factor matrices (line~\ref{alg:proposed:init}).
For each iteration, we update a factor matrix while keeping all others fixed (lines~\ref{alg:proposed:inner_start} to~\ref{alg:proposed:inner_end}).
The time factor matrix is updated with Adam optimizer (line~\ref{alg:proposed:timefactor}) until the validation RMSE increases, which is our convergence criterion (line 8) for Adam.
Each of the non-time factor matrices is updated with the row-wise update rule (line~\ref{alg:proposed:inner_end}) in ALS.
We repeat this process until the validation RMSE continuously increases for five iterations, which is our global convergence criterion (line 12).

\section{Experiment}
\label{sec:experiment}

We perform experiments to answer the following questions.
\begin{itemize}
	\item [\textbf{Q1}] {
		\textbf{Accuracy (Section~\ref{sec:exp:accuracy}).}
		How accurately does \method factorize real-world temporal tensors and predict their missing entries compared to other methods?
	}
	\item [\textbf{Q2}] {
		\textbf{Effect of data sparsity (Section~\ref{sec:exp:sparsity}).}
		How does the sparsity of input tensors affect the predictive performance of \method and other methods?
	}
	\item [\textbf{Q3}] {
		\textbf{Effect of optimization (Section~\ref{sec:exp:optimization}).}
		How effective is our proposed optimization approach for training \method?
	}
	\item [\textbf{Q4}] {
		\textbf{Hyper-parameter study (Section~\ref{sec:exp:hyper}).}
		How do the different hyper-parameter settings affect the performance of \method?
		}
\end{itemize}

\subsection{Experimental Settings}

\subsubsection{Machine}
All experiments are performed on a machine
	equipped with Intel Xeon E5-2630 CPU and a Geforce GTX 1080 Ti GPU.

\subsubsection{Datasets}

\begin{table*}
	\centering
	\vspace{5mm}
	\begin{threeparttable}
		\caption{ \label{tab:dataset}
			Summary of real-world tensors used for experiments.
			Bold text denotes time mode
		}
		\setlength{\tabcolsep}{10pt}
		\begin{tabular}{ l c r r c}
			\toprule
			\textbf{Name} & \textbf{Dimensionality} & \textbf{Nonzero} & \textbf{Granularity} &\textbf{Density}\\
			\midrule
		    \bair\tnote{1} & {\bf 35,064} $\times$ 12 $\times$ 6 & 2,454,305 & 1 hour &0.97 \\
			\mair\tnote{2} & {\bf 2,678 } $\times$ 24 $\times$ 14 & 337,759 & 1 day & 0.37 \\
			\radar\tnote{3} & {\bf 17,937} $\times$ 23 $\times$ 5 & 495,685 & 1 hour &0.24 \\
			\indoor\tnote{4} & {\bf 19,735} $\times$ 9 $\times$ 2 & 241,201 & 10 minutes & 0.70 \\
			\server\tnote{5} & 3 $\times$ 3 $\times$ 34 $\times$ {\bf 4,157}  & 1,009,426 & 1 second & 0.79 \\
	 		\bottomrule
		\end{tabular}
		\scriptsize
		\begin{tablenotes}
			\item[1] {\url{https://archive.ics.uci.edu/ml/datasets/Beijing+Multi-Site+Air-Quality+Data}}
			\item[2] {\url{https://www.kaggle.com/decide-soluciones/air-quality-madrid}}
			\item[3] {\url{https://data.austintexas.gov/Transportation-and-Mobility/Radar-Traffic-Counts/}}
			\item[4] {\url{https://archive.ics.uci.edu/ml/datasets/Appliances+energy+prediction}}
			\item[5] {\url{https://zenodo.org/record/3610078#.XlNpAigzaM8}}
		\end{tablenotes}
	\end{threeparttable}	
\end{table*}

We evaluate \method on five real-world datasets summarized in Table \ref{tab:dataset}.

\begin{itemize}
 	\item \textbf{\bair}~\Citep{zhang2017cautionary}
	is a 3-mode tensor (hour, locations, atmospheric pollutants)
		containing measurements of pollutants.
 	It was collected from $12$ air-quality monitoring sites in Beijing between 2013 to 2017.
	\item \textbf{\mair} is a 3-mode tensor (day, locations, atmospheric pollutants) containing measurements of pollutants in Madrid between 2011 to 2018.
 	\item \textbf{\radar} is a 3-mode tensor (hour, locations, directions) containing traffic volumes measured by radar sensors from 2017 to 2019 in Austin, Texas.
	\item \textbf{\indoor} is a 3-mode tensor (10 minutes, locations, ambient conditions) containing measurements.
		There are two ambient conditions defined: humidity and temperature.
		We construct a fully dense tensor from the original dataset and randomly sample $70$ percent of the elements to make a tensor with missing entries.
		In Section~\ref{sec:exp:sparsity}, we sample from the fully dense version of it.
	\item \textbf{\server} is a 4-mode tensor (air conditioning, server power, locations, second) containing temperatures recorded in a server room.
		The first mode "air conditioning" means air conditioning temperature setups ($24$, $27$, and $30$ Celsius degrees); the second mode "server power" indicates server power usage scenarios ($50\%$, $75\%$, and $100\%$). 
\end{itemize}

Before applying tensor factorization,
we z-normalize the datasets.
Each data is randomly split into training, validation, and test sets with the ratio $8$:$1$:$1$; the validation set is used for determining early stopping.

\subsubsection{Competitors}
We compare \method with the state-of-the-art methods for missing entry prediction.
All the competitors use only the observed entries of a given tensor.
\begin{itemize}
	\item \textbf{\cpals}~\Citep{harshman1970foundations}: a standard CP decomposition method using ALS.
	\item \textbf{CP-WOPT}~\Citep{acar2011scalable}: a CP decomposition method solving a weighted least squares problem.
	\item \textbf{CoSTCo}~\Citep{liu2019costco}: a CNN-based tensor decomposition method.
	\item \textbf{TRMF}~\Citep{yu2016temporal}: a temporally regularized matrix/tensor factorization method.
	\item \textbf{NTF}~\Citep{wu2019neural}: a tensor factorization method integrating LSTM to model time-evolving interactions for rating prediction.
\end{itemize}
	
\subsubsection{Metrics}
We evaluate the performance using RMSE (Root Mean Squared Error) and MAE (Mean Absolute Error) defined as follows.
\begin{align*}
	\text{RMSE} = {\sqrt{\frac{1}{|\Omega|}\sum_{\forall\alpha \in \Omega}{\left(x_{\alpha}-\hat{x}_{\alpha}\right)^{2}}}},
	\quad
	\text{MAE} = \frac{1}{|\Omega|}{\sum_{\forall\alpha \in \Omega}{|x_{\alpha}-\hat{x}_{\alpha}|}}
\end{align*}
$\Omega$ indicates the set of the indices of observed entries.
$x_\alpha$ stands for the entry with index $\alpha$ and
	$\hat{x}_{\alpha}$ is the corresponding reconstructed value.

\subsubsection{Hyper-parameter}
We use hyper-parameters in Table~\ref{tab:hyper-param} for \method,
	except in Section~\ref{sec:exp:hyper} where we vary hyper-parameters.
We use $0.5$ for $\sigma$ which adjusts the smoothing level in kernel function.
We change the window size $ S \in \{3, 5, 7, 9, 11 \}$ and find the optimal value for each dataset.		
\begin{table}[t!]
	\small
	\centering
	\caption{Default hyper-parameter setting}
 	\setlength{\tabcolsep}{4pt}
	\begin{tabular}{ l c c c c }
		\toprule
		\textbf{Dataset} & \textbf{{\makecell{Learning \\ rate $\eta$ }}} &
		\textbf{Rank $K$} &
		\textbf{Window $S$} & \textbf{Penalty $\lambda_t$} \\
		\midrule
		\bair   & $10^{-2}$ & $10$ & $3$ & $10^{3}$ \\
		\mair   & $10^{-2}$ & $10$ & $5$ & $10^{2}$ \\
		\radar  & $10^{-2}$ & $10$ & $9$ & $10^{2}$ \\
		\indoor & $10^{-2}$ & $10$ & $3$ & $10^{2}$ \\
		\server & $10^{-3}$ & $10$ & $3$ & $10^{-1}$ \\
	\bottomrule
	\end{tabular}	
	\label{tab:hyper-param}
\end{table}


\subsection{Accuracy (Q1)} \label{sec:exp:accuracy}

We compare \method with competitors in terms of RMSE and MAE in Table~\ref{tab:standarderror}.
\methodz indicates \method without the sparsity penalty.
Note that \method consistently gives the best accuracy for all the datasets.
\method achieves up to $7.01\times$ lower RMSE
	and $5.50\times$ lower MAE compared to the second-best methods.
 \methodz provides the second-best performance; the smoothing regularization effectively predicts missing values by capturing temporal patterns and leaving out the noise.


\begin{table*}
	\centering
	\vspace{5mm}
	\caption{
		Performance of missing entry prediction by \method and competitors.
		The best method is in bold, and the second-best method is underlined.
		Our proposed \method consistently shows the best performance in all datasets
	}	
	\renewcommand\arraystretch{1.2}
	\setlength{\tabcolsep}{4.5pt} 	
	\begin{tabular}{l | c | c | c | c | c}
		\toprule
		Data &
		{\makecell{Beijing \\ Air quality}}  &
		{\makecell{Madrid \\ Air quality}}  &
		{\makecell{Radar \\ Traffic}}  &
		{\makecell{Indoor \\ Condition}}  &
		{\makecell{Server \\ Room}} \\
	    \midrule
		\diagbox[height=2em]{Method}{Metric}
		& RMSE / MAE & RMSE / MAE
		& RMSE / MAE & RMSE / MAE & RMSE / MAE
		\\ \midrule
		\cpals
		& 0.352  / 0.219 & 0.456 / 0.293
		& 0.365  / 0.248 & 0.624 / 0.316
		& 0.076 / 0.048
		\\ [0.3em]
		CP-WOPT
		& 0.766 /  0.538 &  0.482 /  0.297
		&  0.328 /  0.206 &  0.603 /  0.310
		&  0.070 /  0.046
		\\ [0.3em]
		CoSTCo
		&  0.360 /  0.223 &  0.461 / 0.303
		&  0.298 /  0.197 &  0.609 / 0.303
		&  0.306 /  0.090
		\\ [0.3em]
		TRMF
		&  1.098 / 0.770 & 1.004 / 0.936
		& 0.695 / 0.485 & 0.894 / 0.563 & 1.083 / 0.813
		\\ [0.3em]
		NTF
		&  0.529 /  0.333 &  0.648 /  0.455
		&  0.585 /  0.400 &  0.968 /  0.576  & 0.660 / 0.516
		\\ [0.3em]
		\midrule
		\methodz
		& \underline{0.327} / \underline{0.204}
		& \underline{0.416} / \underline{0.279}
		& \underline{0.257} / \underline{0.160}
		& \underline{0.088} / \underline{0.057}
		& \underline{0.058} / \underline{0.039}
		\\ [0.3em]
		\textbf{\method (proposed)}
		& \textbf{0.323} / \textbf{0.201}
		& \textbf{0.409} / \textbf{0.274}
		& \textbf{0.249} / \textbf{0.152}
		& \textbf{0.086} / \textbf{0.055}
		& \textbf{0.054} / \textbf{0.035}
		\\
	\bottomrule
	\end{tabular}
	\label{tab:standarderror}
\end{table*}

\subsection{Effect of Data Sparsity (Q2)} \label{sec:exp:sparsity}
\begin{figure*}
	\addvspace{6mm}
	\centering
    \includegraphics[width=0.75\linewidth]{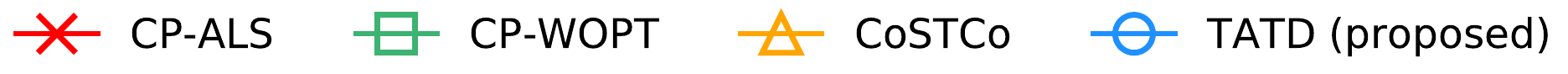}\vspace{-1mm}\\
    \subfigure[\bair]{
        \includegraphics[width=0.32\linewidth]{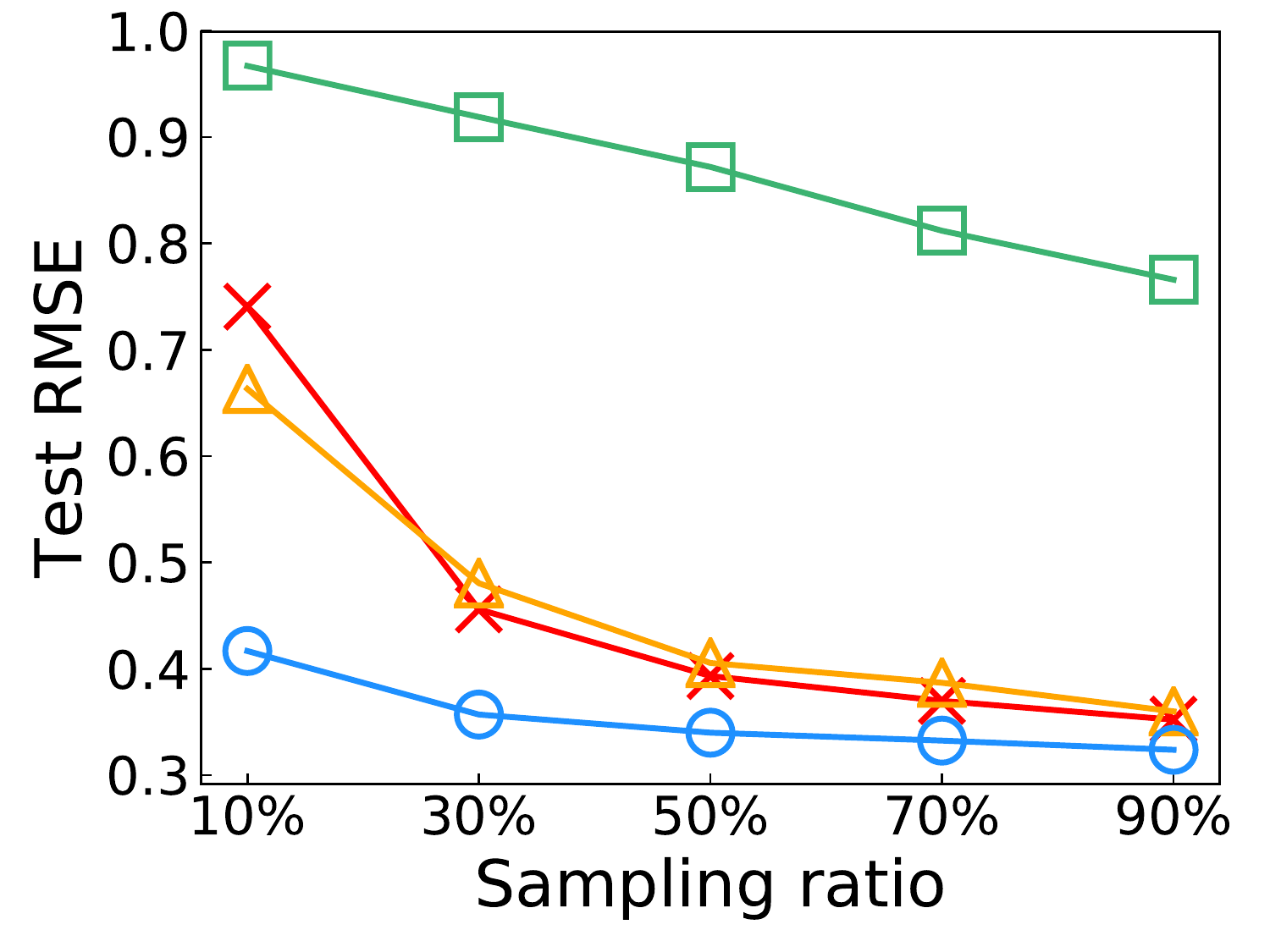}
    }
    \hspace{-2mm}
    \subfigure[\mair]{
        \includegraphics[width=0.32\linewidth]{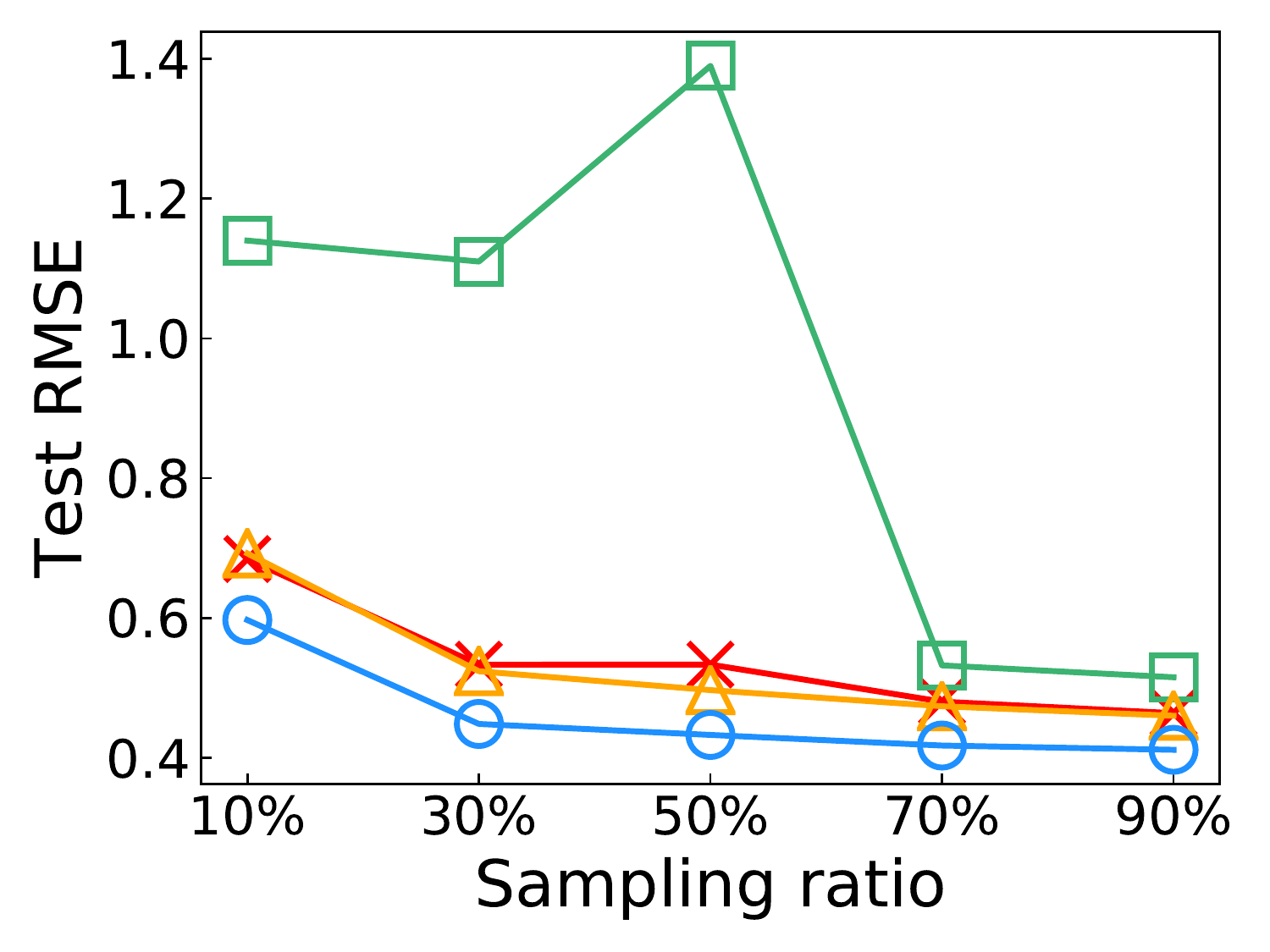}
    }
    \hspace{-2mm}
    \subfigure[ \radar]{
        \includegraphics[width=0.32\linewidth]{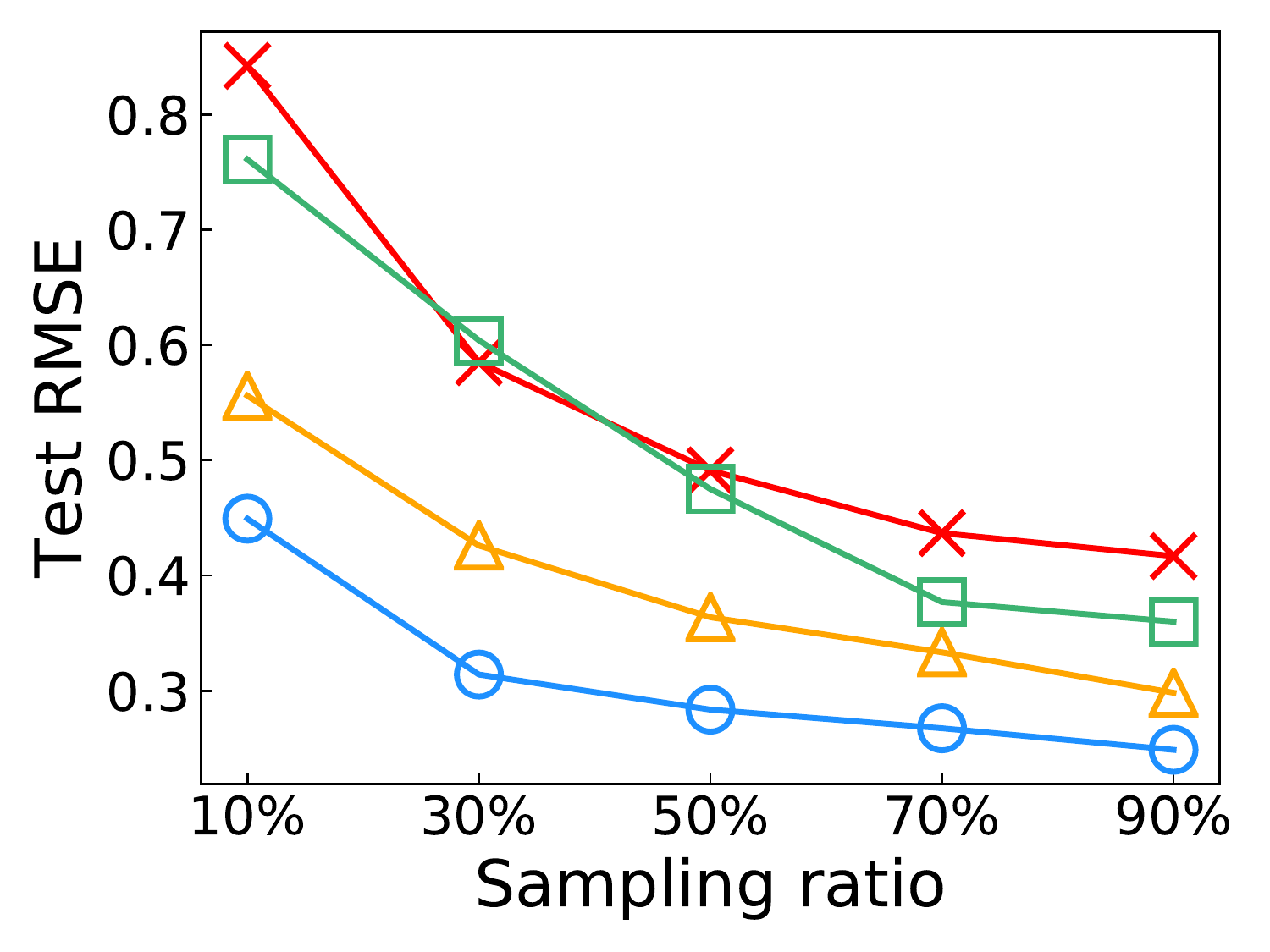}
    }\\
    \subfigure[ \indoor]{
        \includegraphics[width=0.32\linewidth]{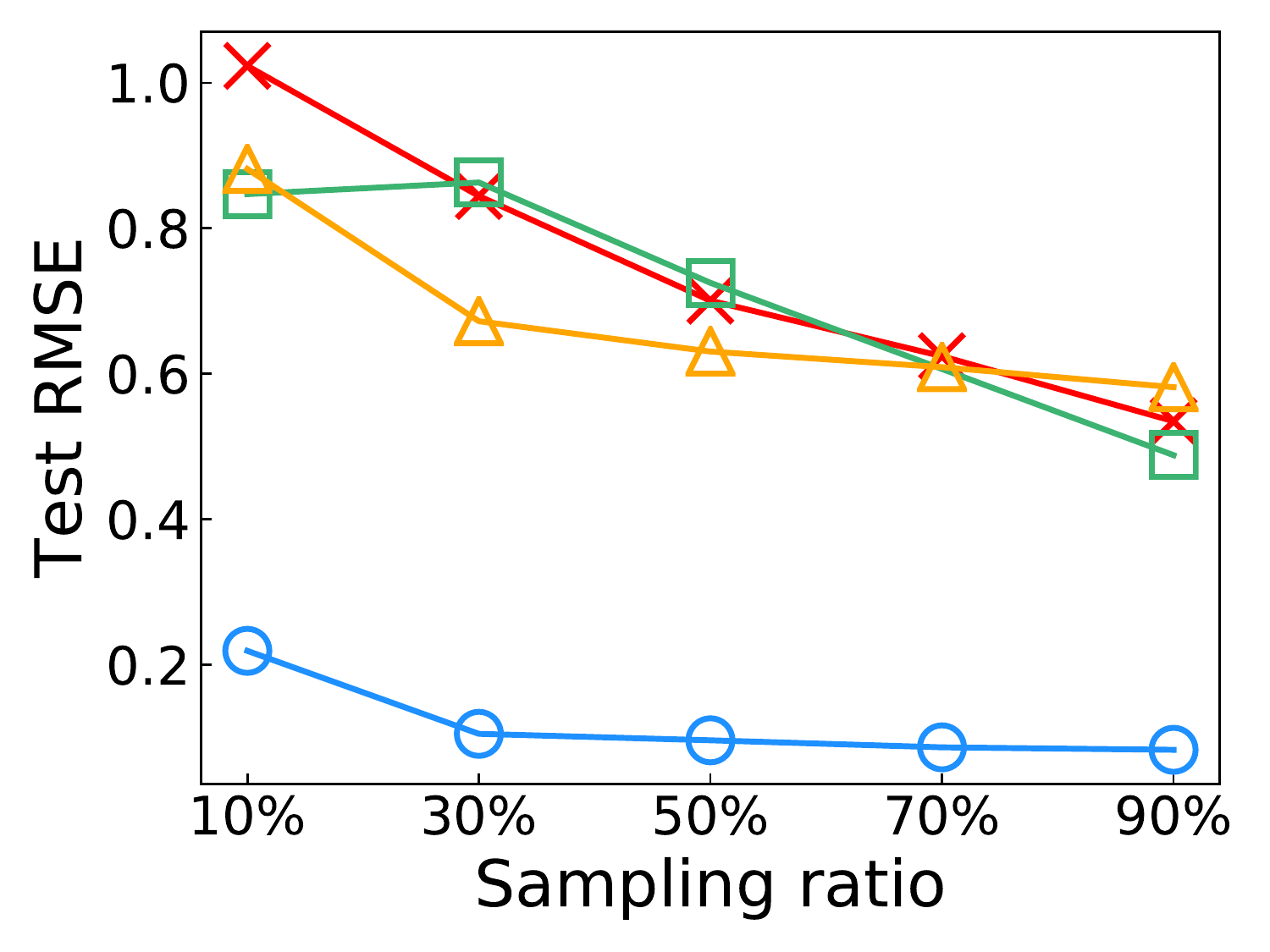}
    }
    \hspace{-2mm}
    \subfigure[ \server]{
        \includegraphics[width=0.32\linewidth]{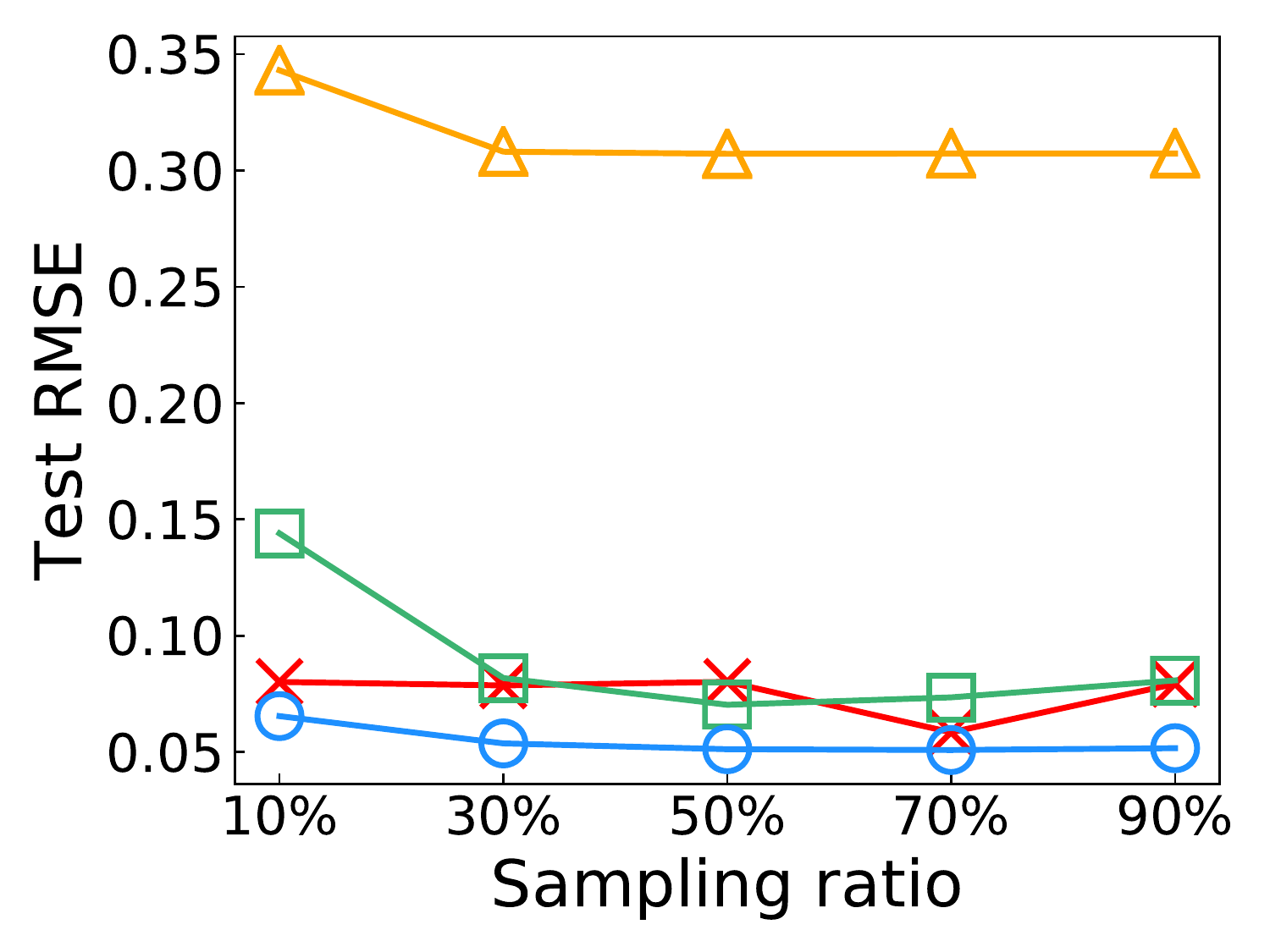}
    }
    \caption{\label{fig:sparsity_ratio}
	    Test RMSE of \method and the best competitors CP-ALS, CP-WOPT and CoSTCo for varying data sampling ratio.
		Note that the error gap of \method
		and competitors becomes larger when the sparsity increases 
		(the sampling ratio decreases),
		due to the careful consideration of sparse time slices by \method. 
		}
\end{figure*}

We evaluate the performance of \method with varying data sparsity.
We sample the data with the ratio of $\{10, 30, 50, 70, 90\}\%$ to identify how accurately the method predicts missing entries even when the data are highly sparse.
Fig.~\ref{fig:sparsity_ratio} shows the errors of \method and the best competitors,
	CP-ALS, CP-WOPT, and CoSTCo, for five datasets.
Note that the error gap of \method and competitors becomes larger
	when the sparsity increases (the sampling ratio decreases).
\method achieves up to $4.67\times$, $3.86\times$ and $5.24\times$ lower test RMSE
		than CP-ALS, CP-WOPT, and CoSTCo, respectively,
		when we use only $10\%$ of data.
There are two reasons for the superior performance of \method as the sparsity increases.
First, \method is designed to infer missing entries of a target slice by using its neighboring slices; this is especially useful when the target slice is extremely sparse and has no information to infer its entries.
Second, \method explicitly considers sparsity in its model through the sparsity penalty,
and imposes more regularization for sparser slices.

\subsection{Effect of Optimization (Q3)} \label{sec:exp:optimization}

We evaluate our optimization strategy in terms of error and running time.
We call our strategy as \alsa and
	compare it with the following optimization strategies.
\begin{itemize}
	\item Adam: a recent gradient-based method using momentum and controlling learning rate.
	\item SGD: a standard stochastic gradient descent method which is widely used for optimization.
	\item ALS + SGD: an alternating minimization method which updates a time factor matrix with SGD and non-time factor matrices with the least square solution.
	\item Alternating Adam: an alternating minimization method
		which updates a single factor matrix with Adam while fixing other factor matrices.
\end{itemize}

\begin{figure*}[t]
	\centering
	\vspace{1mm}
	\includegraphics[width=0.95\textwidth]{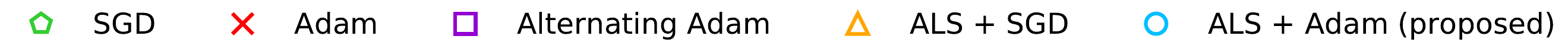}
	\\ \vspace{-1mm}
	\hspace{-3mm}
	\subfigure[ \bair]{
		\includegraphics[width=0.32\textwidth]{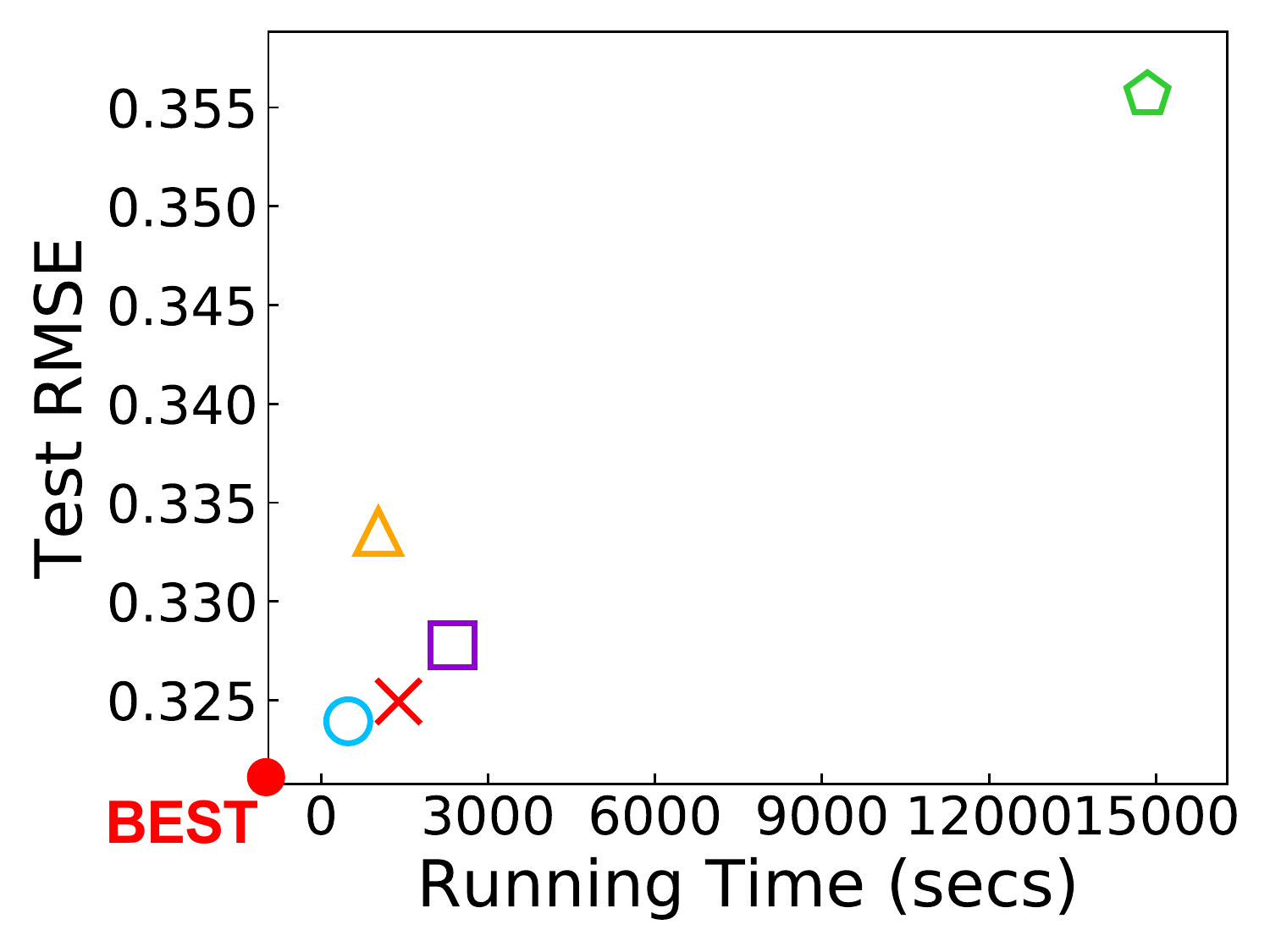}
	}
	\hspace{-3mm}
	\subfigure[ \mair]{
		\includegraphics[width=0.32\textwidth]{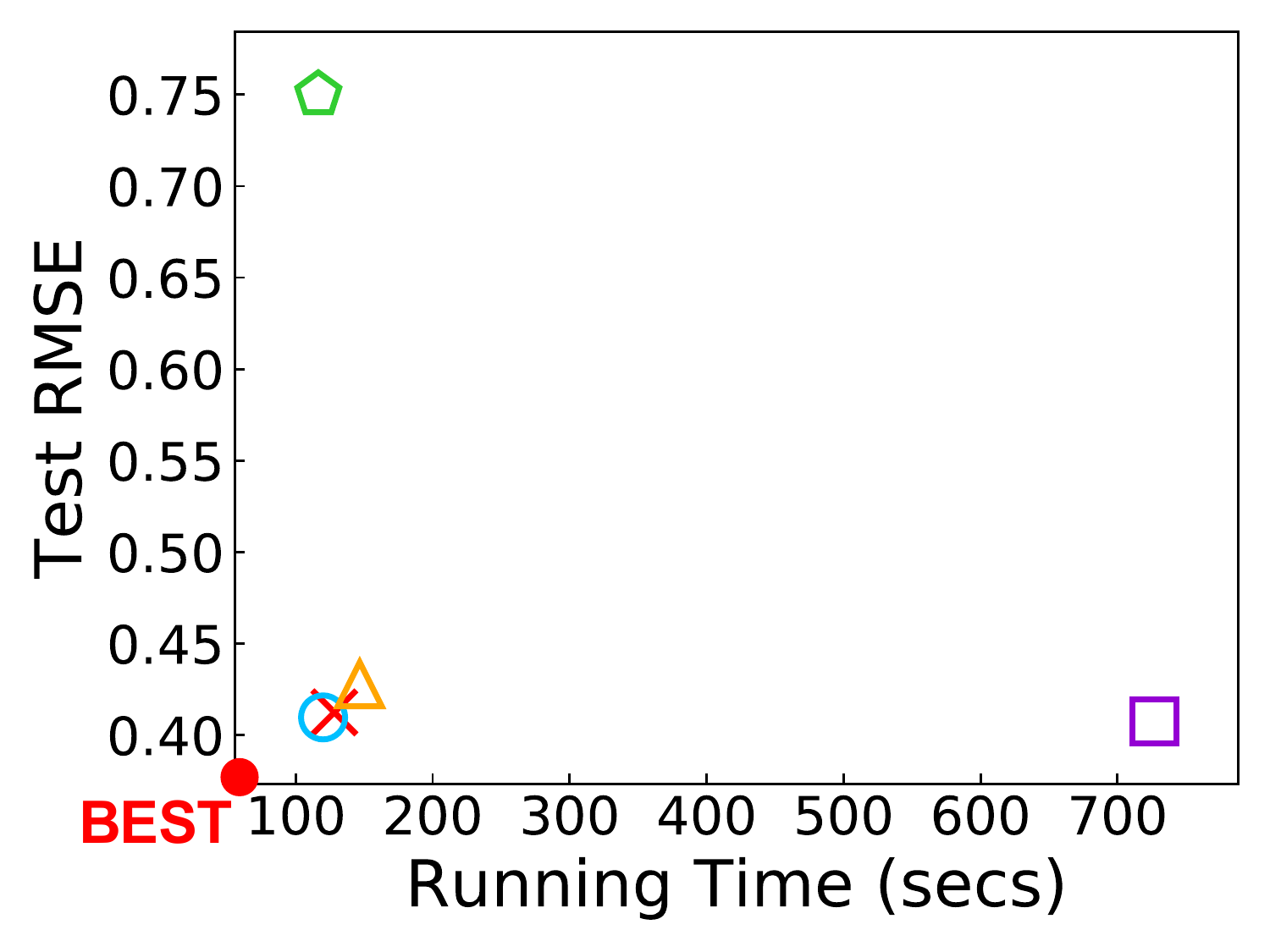}
	}
	\hspace{-3mm}
	\subfigure[ \radar]{
		\includegraphics[width=0.32\textwidth]{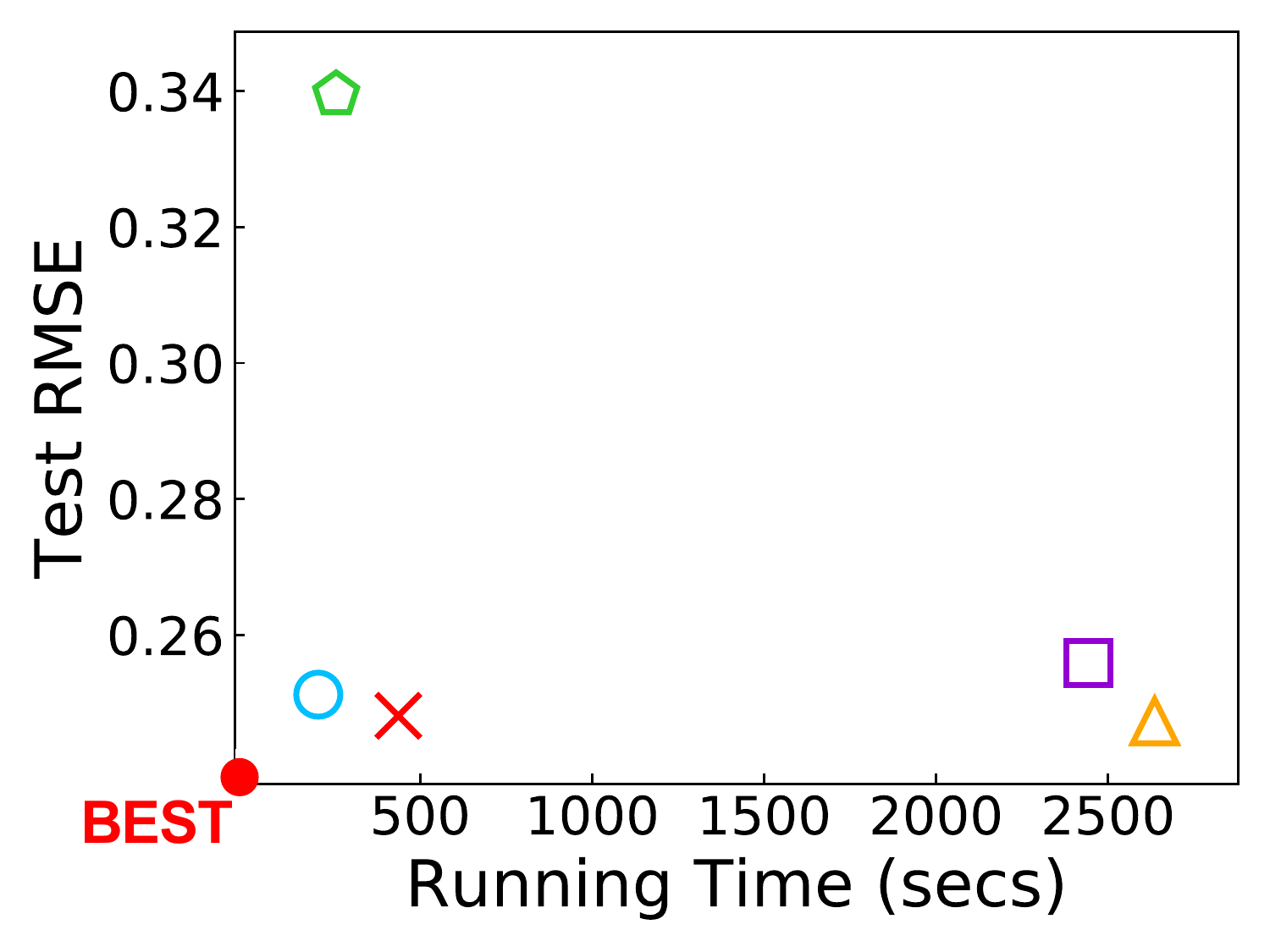}
	}\\
	\hspace{-3mm}
	\subfigure[ \indoor]{
		\includegraphics[width=0.32\textwidth]{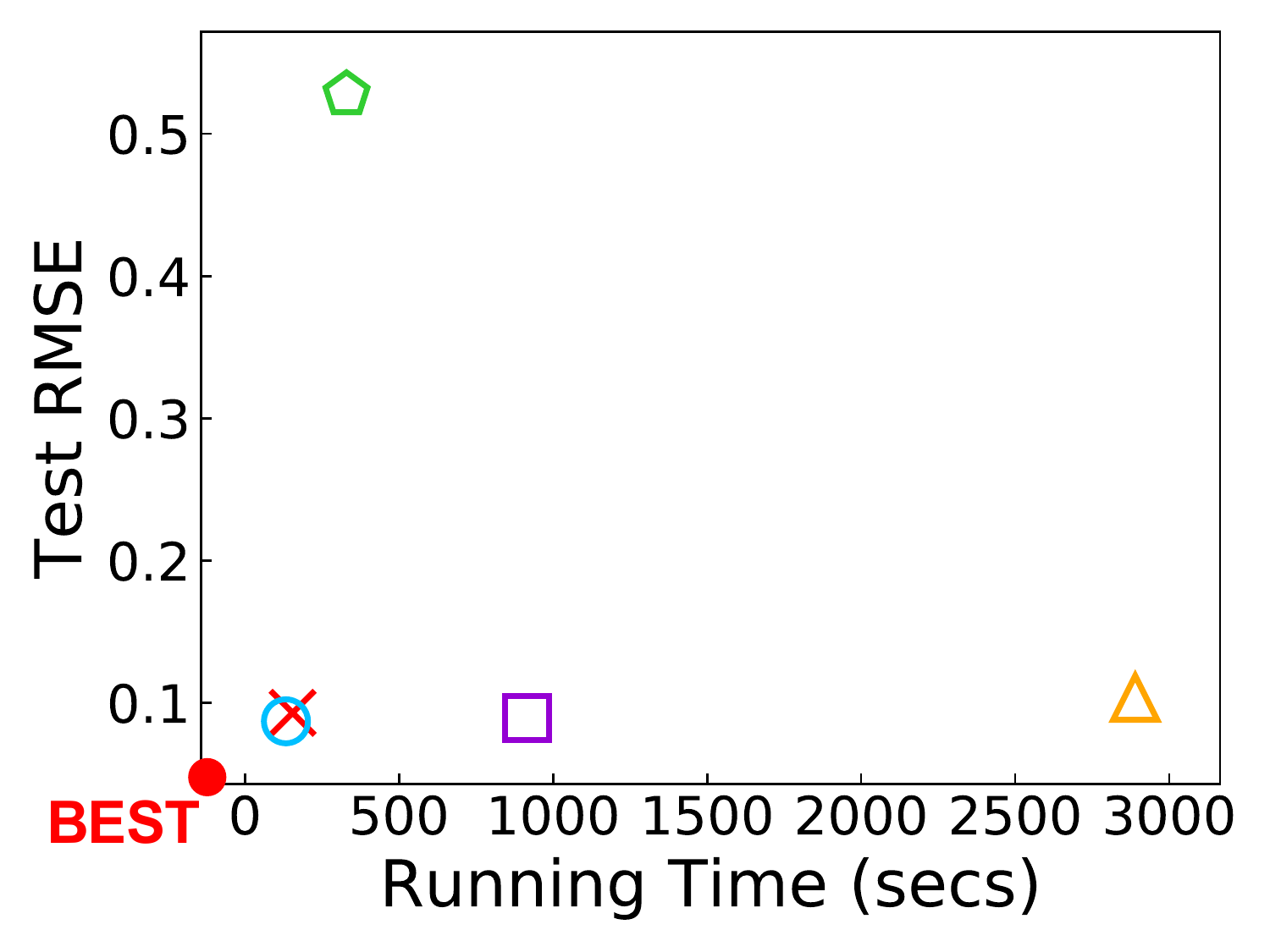}
	}
	\hspace{-3mm}
	\subfigure[ \server]{
		\includegraphics[width=0.32\textwidth]{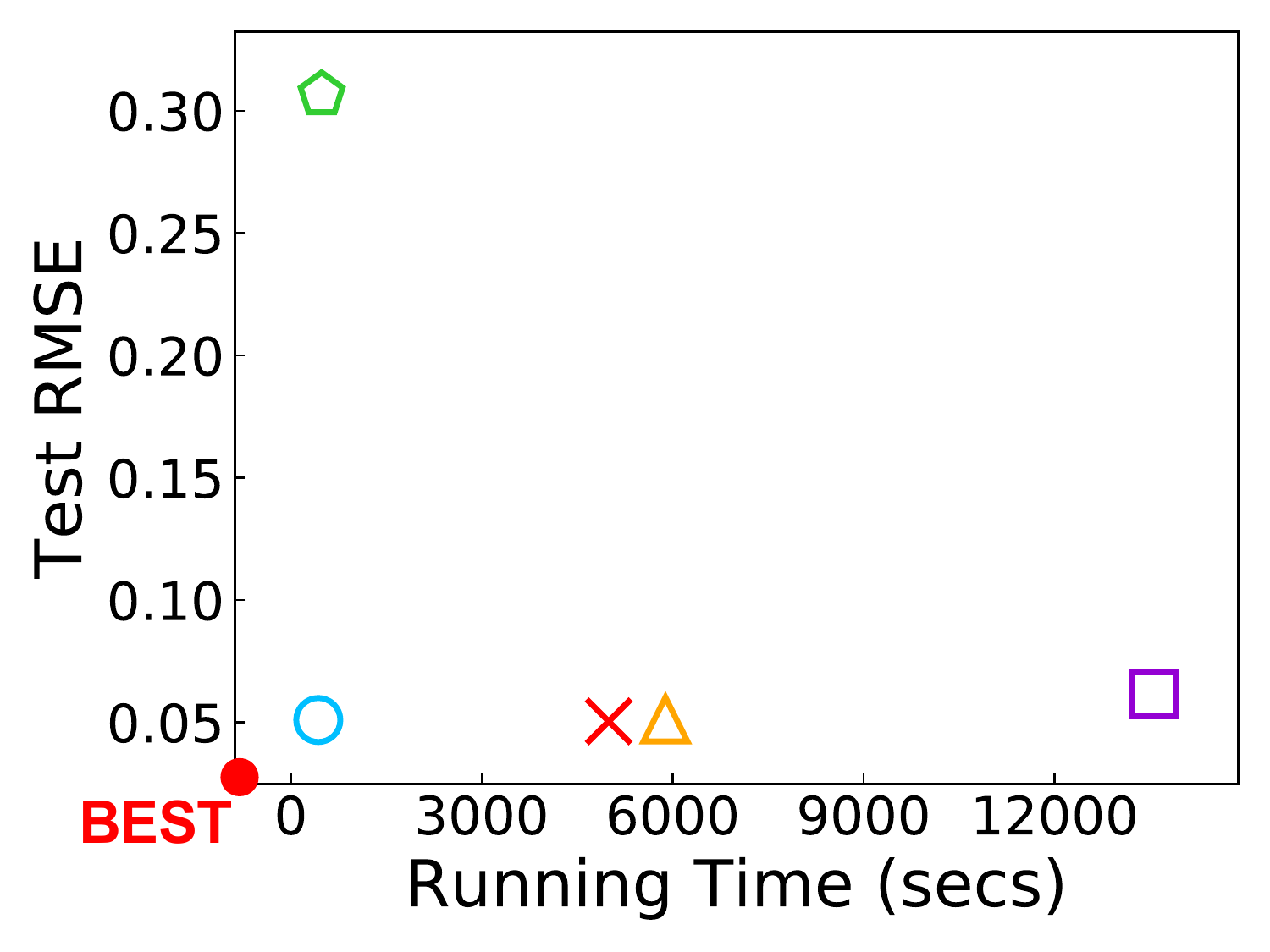}
	}\\
	\caption{\label{fig:opt}	
		Comparison of optimization strategies in \method.
		Our proposed ALS + Adam makes the best trade-off of running time and test RMSE,
			giving smaller running time and test RMSE compared to other methods in general
	}
\end{figure*}

Fig.~\ref{fig:opt} shows the result.
Note that our proposed \alsa makes the best trade-off of running time and test RMSE,
giving smaller running time and test RMSE compared to other methods in general.
\alsa achieves better results compared to ALS + SGD since Adam optimizer finds a better local minimum compared to SGD.
Compared to alternating Adam, \alsa achieves better results as well since updating each non-time factor matrix has an analytical solution by ALS, and thus a gradient-based approach Adam is less effective.
%

\subsection{Hyper-parameter Study (Q4)} \label{sec:exp:hyper}
We evaluate the performance of \method with regard to hyper-parameters:
	smoothing regularization penalty and rank size.
	
\subsubsection{Smoothing regularization penalty}\label{sec:exp:penalty}

\begin{figure*}[t]
	\centering
	\vspace{10mm}
	\subfigure[\bair]{
		\includegraphics[width=0.3\linewidth]{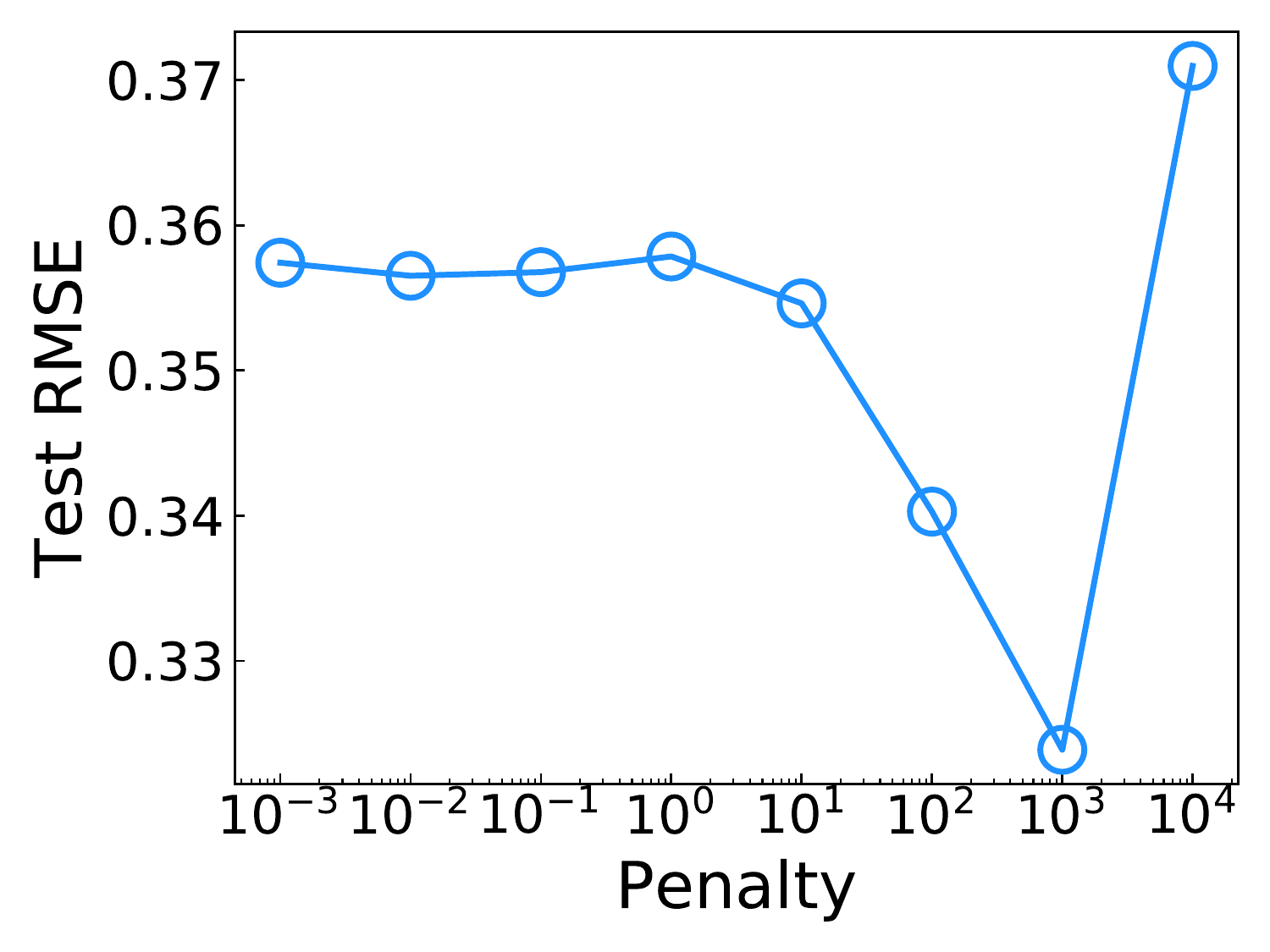}
	}
	\hspace{-2mm}
	\subfigure[\mair]{
		\includegraphics[width=0.3\linewidth]{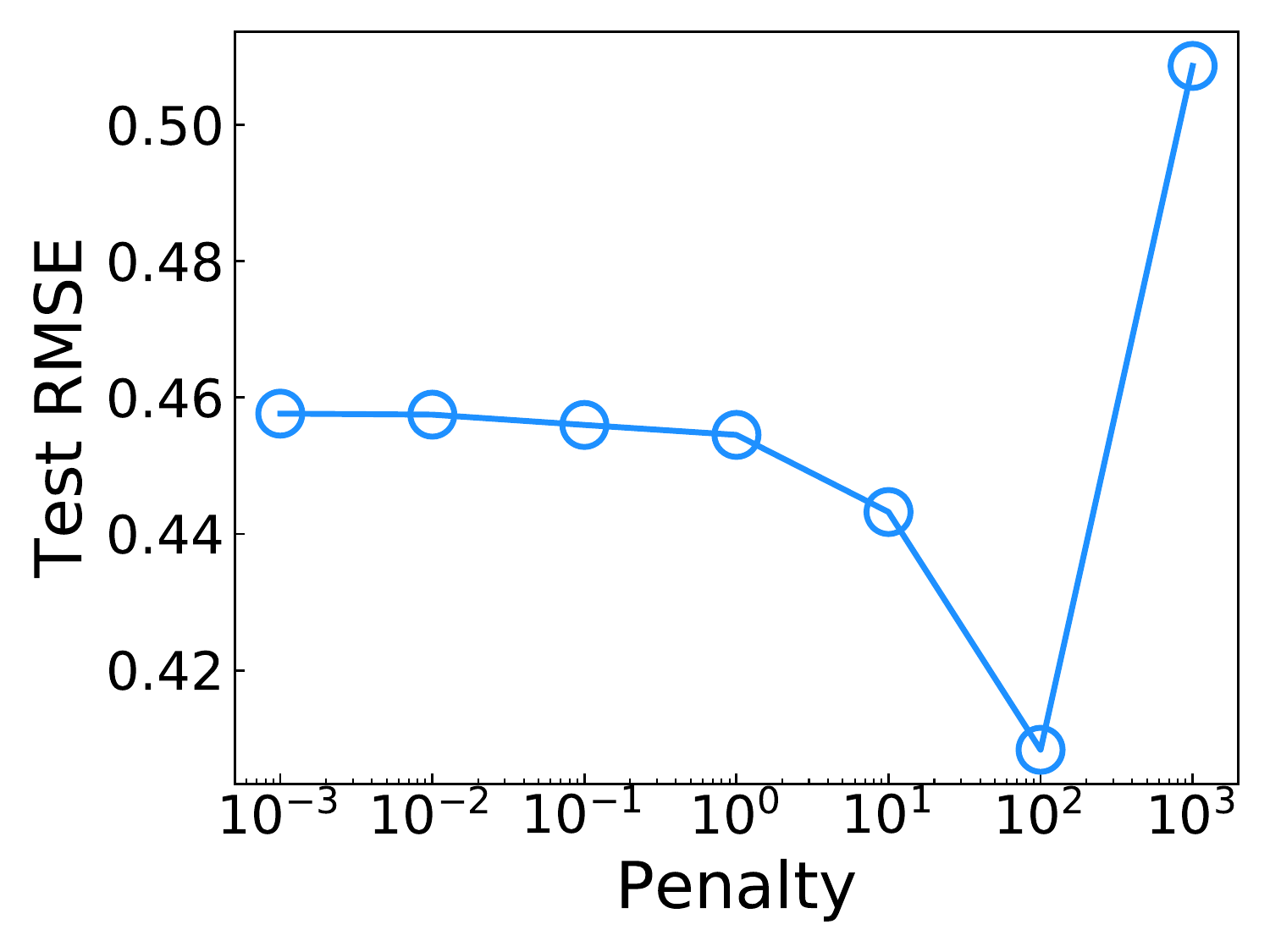}
	}
	\hspace{-2mm}
	\subfigure[\radar]{
		\includegraphics[width=0.3\linewidth]{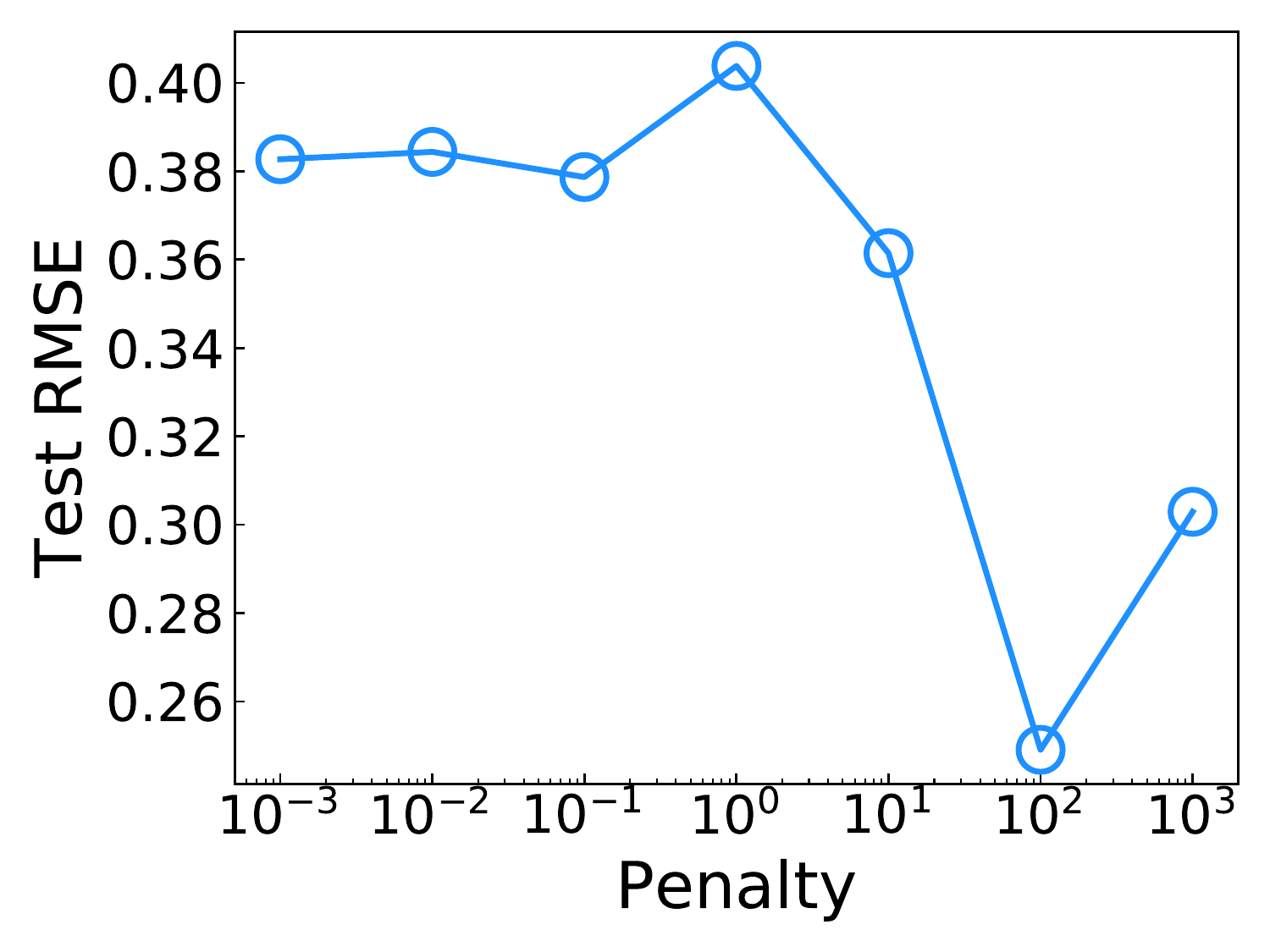}
	}\\
	\subfigure[\indoor]{
		\includegraphics[width=0.3\linewidth]{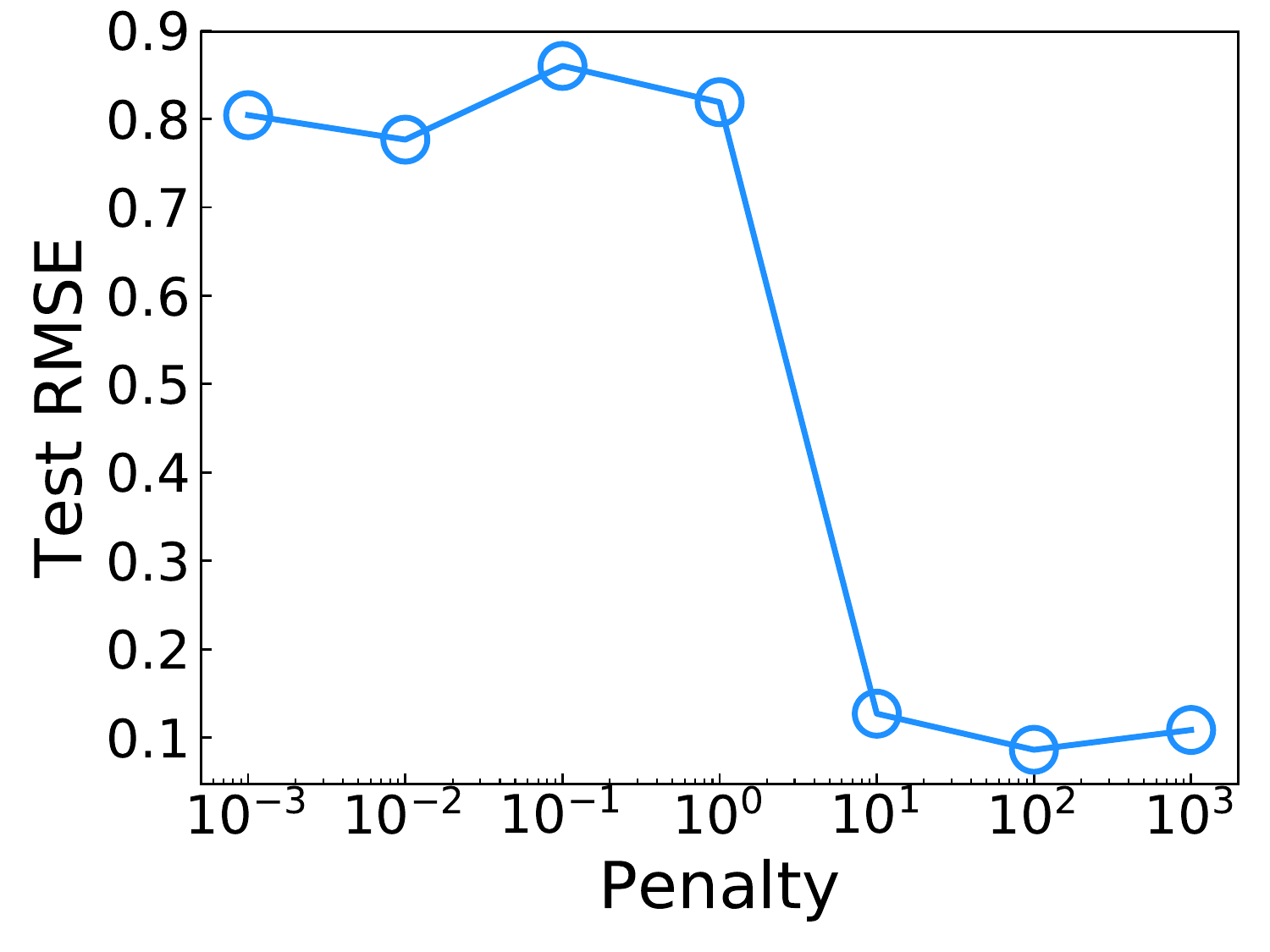}
	}
	\hspace{-2mm}
	\subfigure[\server]{
		\includegraphics[width=0.3\linewidth]{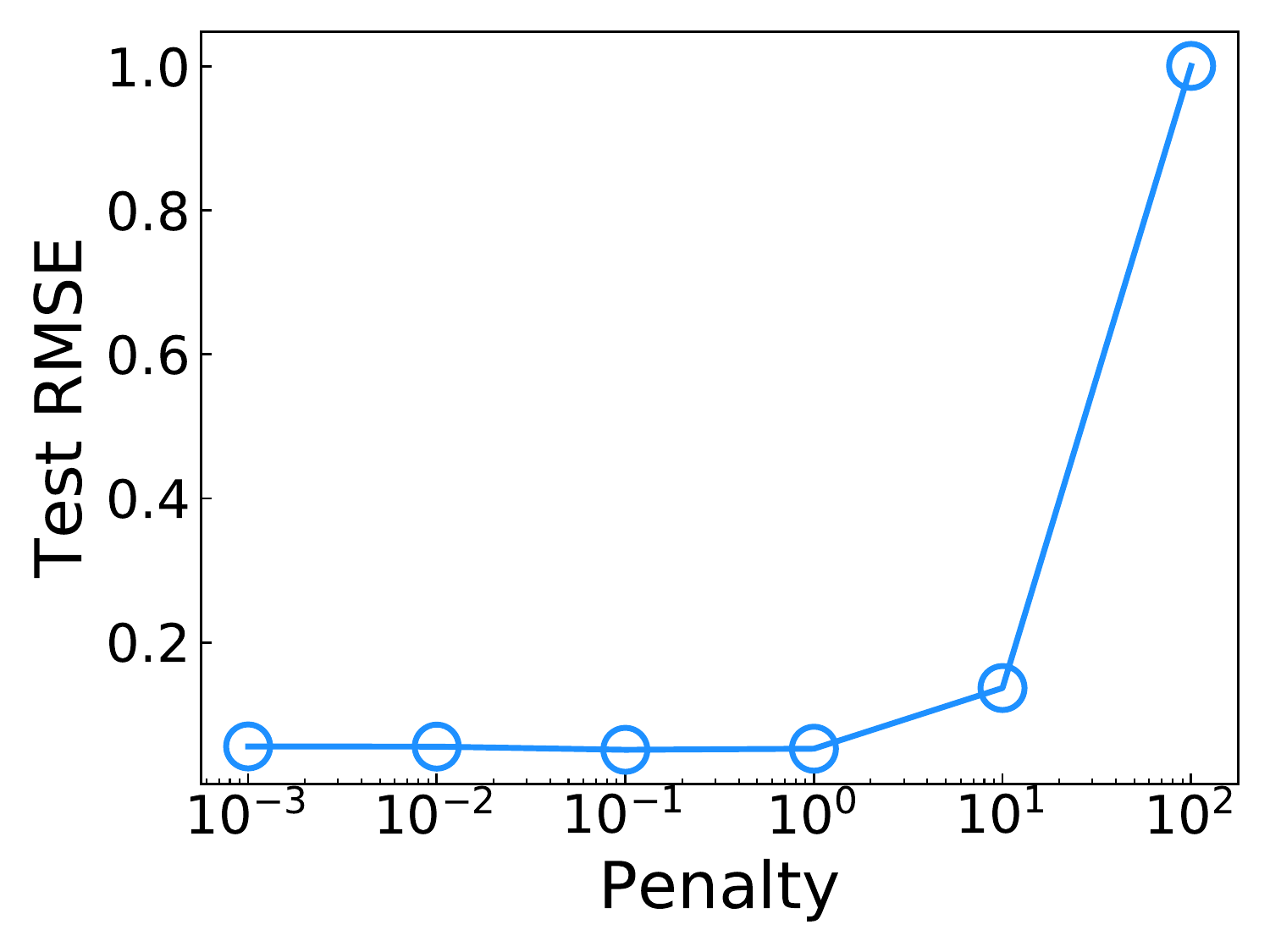}
	}\\
	\caption{ \label{fig:penalty}
		Effect of the smoothing regularization penalty parameter $\lambda_t$ in \method.
		Note that too small or too large values of $\lambda_t$ lead to overfitting and underfitting, respectively.
		A right amount of smoothing regularization gives the smallest error,
			verifying the effectiveness of our proposed idea
	}
\end{figure*}

We vary the smoothing regularization penalty $\lambda_t$
	and evaluate the test RMSE in Fig.~\ref{fig:penalty}.
Note that too small or too large values of $\lambda_t$ do not give the best results;
too small value of $\lambda_t$ leads to overfitting, and too large value of it leads to underfitting.
The results show that a right amount of smoothing regularization gives the smallest error, verifying the effectiveness of our proposed idea.


\subsubsection{Rank}\label{sec:exp:rank}
\begin{figure*}
	\centering
	\includegraphics[width=0.75\linewidth]{./fig/legend/four_legend}\\
	\subfigure[\bair]{
		\includegraphics[width=0.3\linewidth]{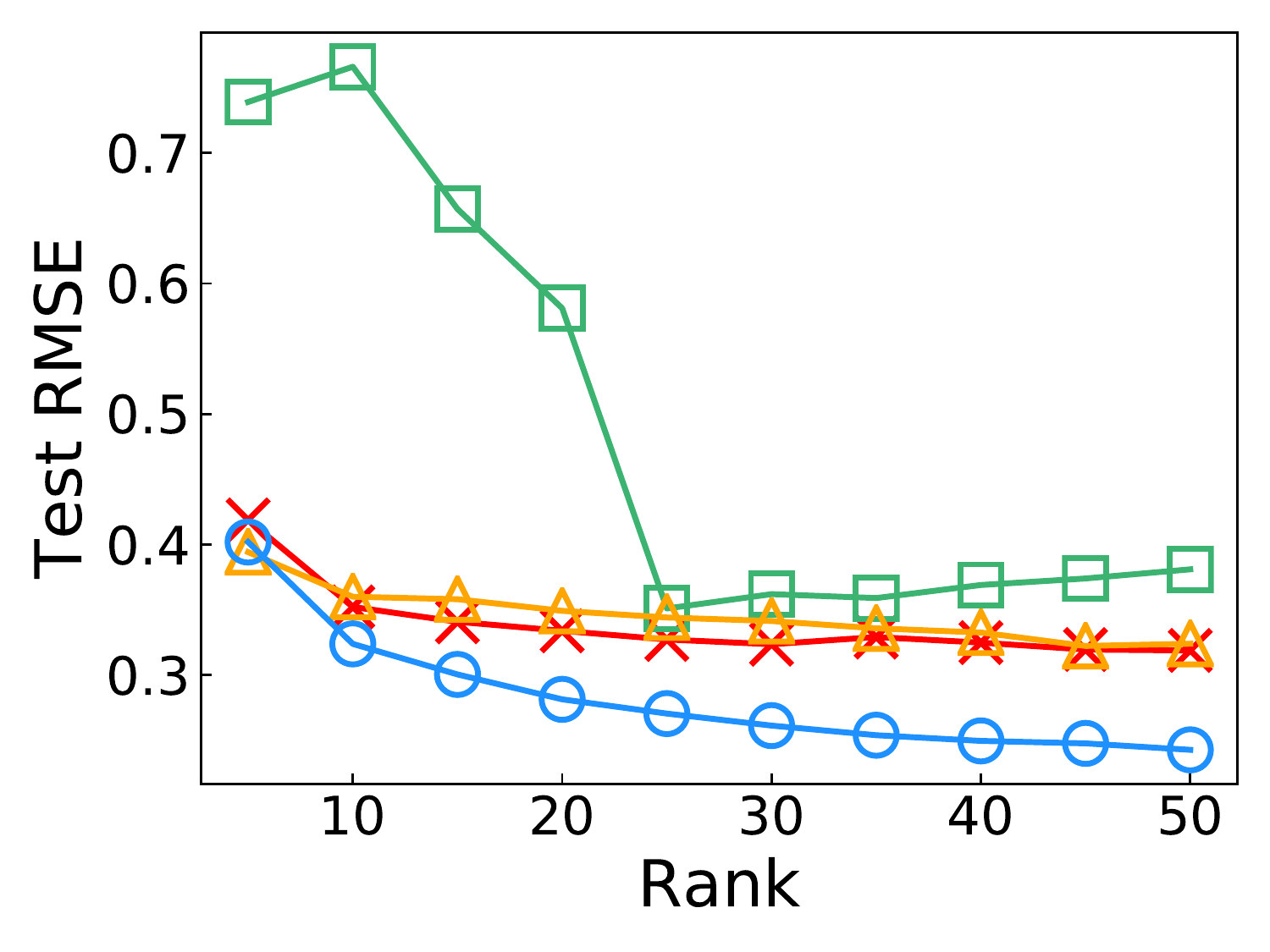}
	}
	\hspace{-2mm}
	\subfigure[\mair]{
		\includegraphics[width=0.3\linewidth]{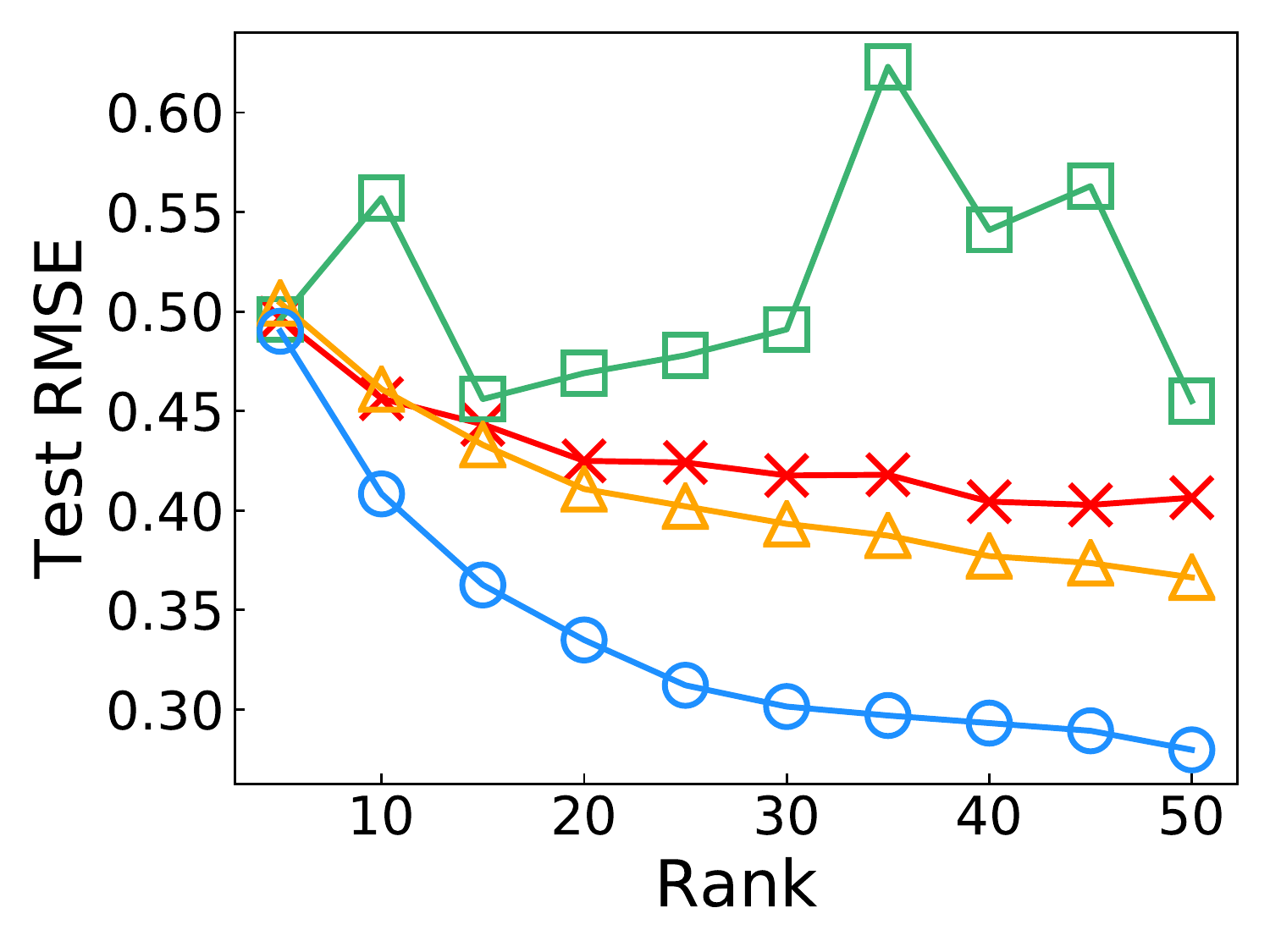}
	}
	\hspace{-2mm}
	\subfigure[\radar]{
		\includegraphics[width=0.3\linewidth]{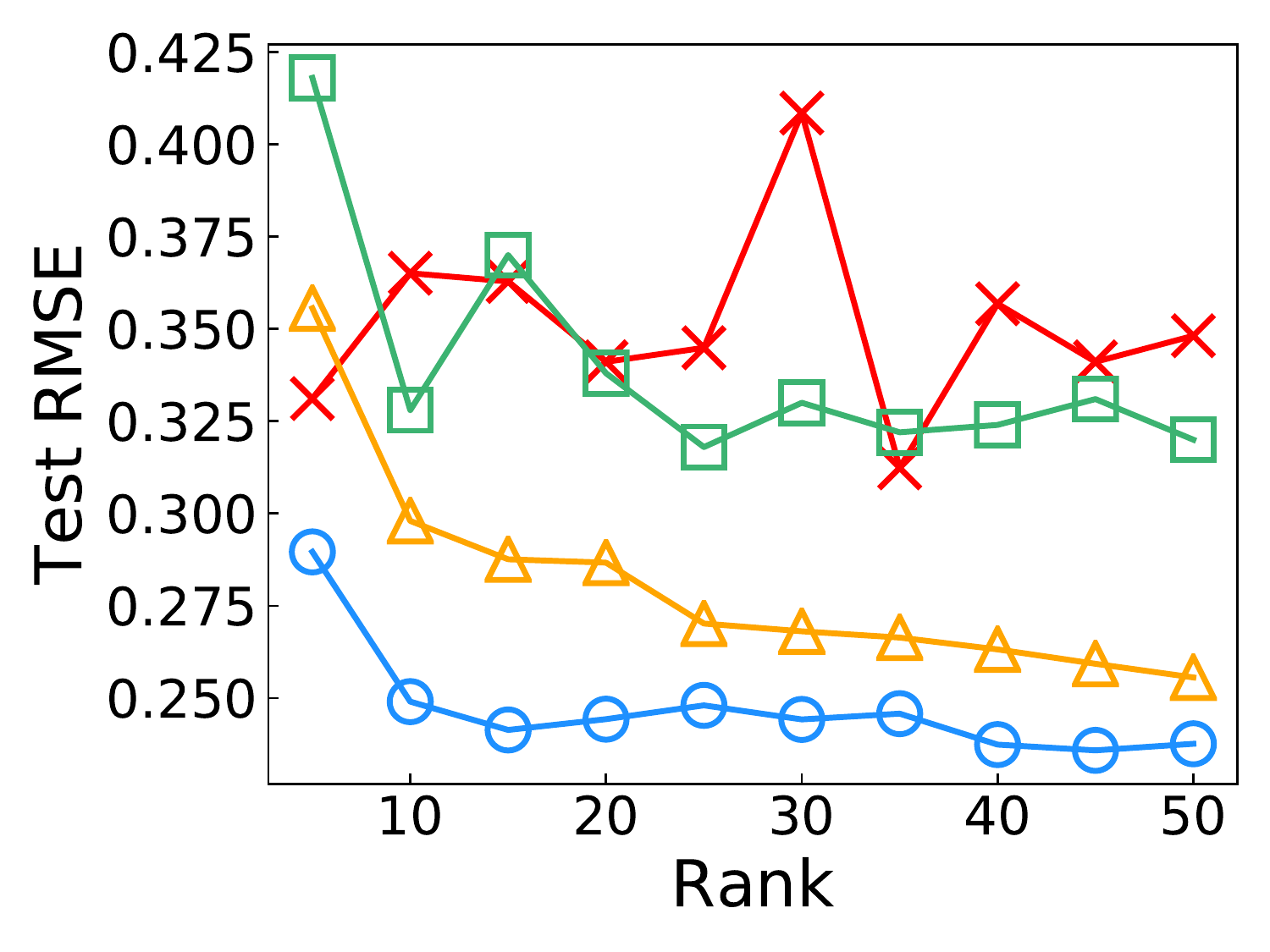}
		\label{fig:rank_radar}
	}\\
	\subfigure[\indoor]{
		\includegraphics[width=0.3\linewidth]{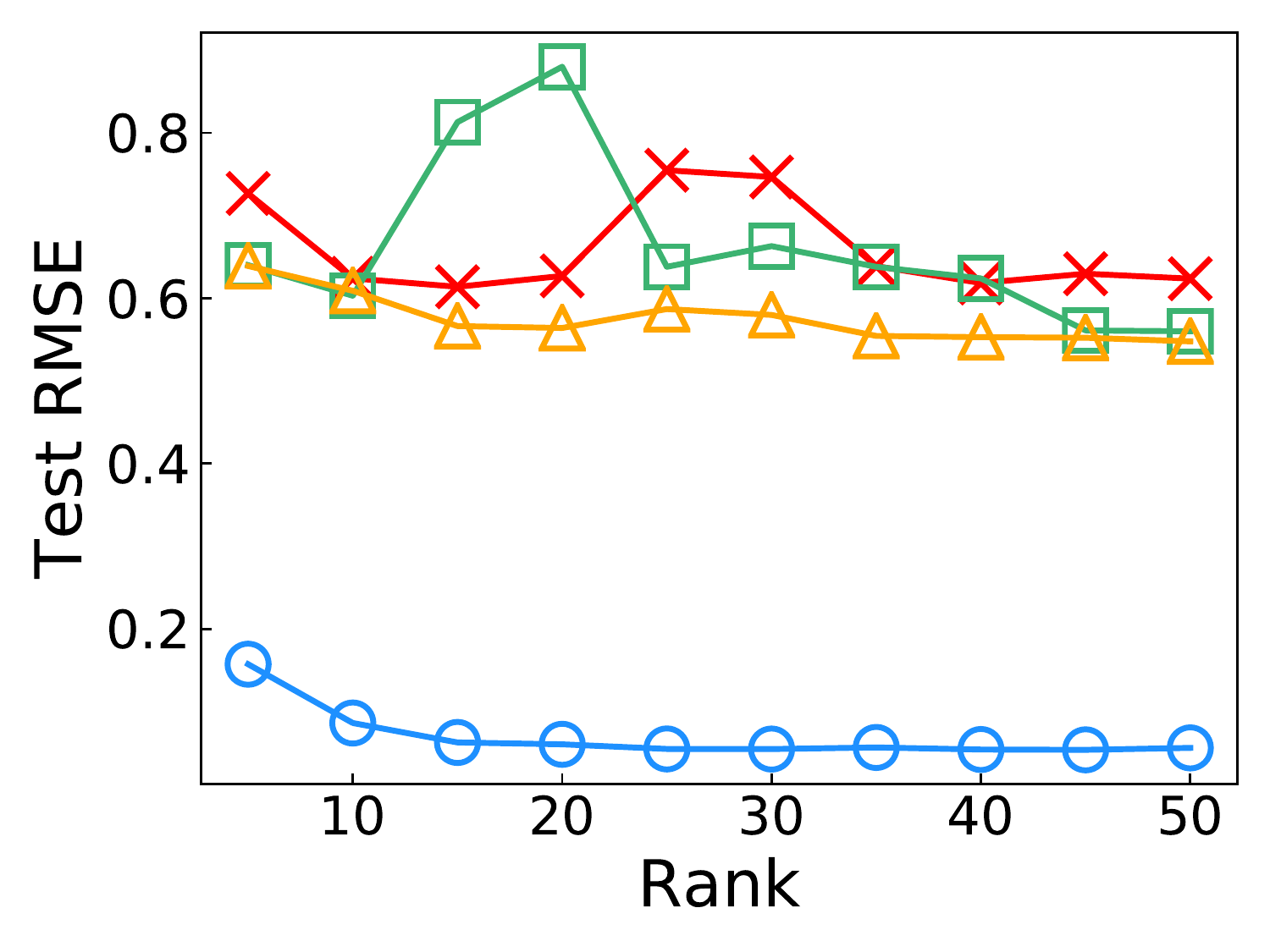}
		\label{fig:rank_indoor}
	}
	\hspace{-2mm}
	\subfigure[\server]{
		\includegraphics[width=0.3\linewidth]{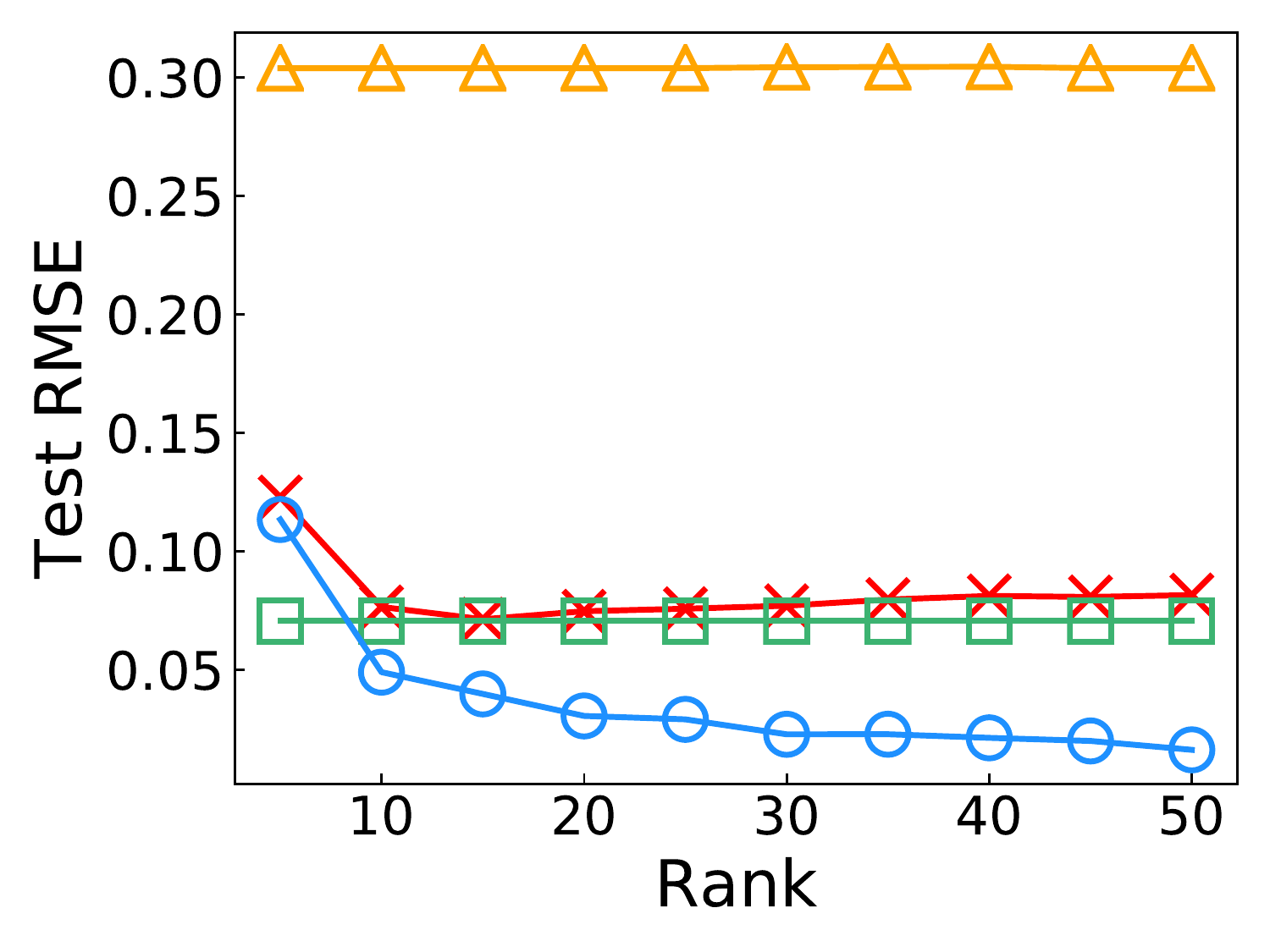}
		\label{fig:rank_server}
	}\\
	
	\caption{\label{fig:rank}
	Effect of rank on the performance of \method.
	When the rank increase,
	\method shows a stable performance improvement unlike competitors,
	and
	the error gap between \method and competitors increases.
	\method works even better for higher ranks since it exploits rich information from neighboring rows when regularizing a row of the time factor matrix.
	}
\end{figure*}

We increase the rank $K$ from $5$ to $50$ and evaluate the test RMSE in Fig.~\ref{fig:rank}.
We have two main observations.
First, \method shows a stable performance improvement with increasing ranks,
	compared to CP-ALS and CP-WOPT which show unstable performances.
Second, the error gap between \method and competitors increases with increasing ranks.
Higher ranks may make the models overfit to a training dataset; 
however, 
\method works even better for higher ranks since it exploits rich information from neighboring rows when regularizing a row of the time factor matrix.
%


\section{Related Work}
\label{sec:related}

We describe related works on tensor factorization and missing entry prediction on temporal tensors.

\subsection{Tensor Decomposition}
We present one of the major tensor decomposition methods, CP decomposition.

%
	\textbf{CP decomposition. }
	CP decomposition methods~\Citep{KangPHF12,jeon2015haten2,ChoiV14} have been widely used for analyzing large-scale real-world tensors.
	\cite{KangPHF12,jeon2015haten2} propose distributed CP decomposition methods running on the MapReduce framework.
	\citet{ChoiV14} propose a scalable CP decomposition method by exploiting properties of a tensor operation used in CP decomposition.
	\citet{BattaglinoBK18} propose a randomized CP decomposition method which reduces the overhead of computation and memory.
	However, the above methods are not appropriate for missing value prediction in highly sparse tensors since they assume the values of missing entries are zero.

	Several CP decomposition methods have been developed to handle sparse tensors without setting the values of missing entries as zero.
	\citet{PapalexakisFS12} propose ParCube to obtain sparse factor matrices using a sampling technique in parallel systems.
	\citet{BeutelTKFPX14} propose FlexiFaCT, which performs a coupled matrix-tensor factorization using Stochastic Gradient Descent (SGD) update rules.
	\citet{shin2016fully} propose CDTF and SALS, which are scalable CP decomposition methods for sparse tensors.
	\citet{SmithK17} improves the efficiency of CP decomposition for sparse tensors by exploiting a compressed data structure.
	The above CP decomposition methods do not consider temporal dependency and time-varying sparsity which are crucial for temporal tensors.
On the other hand, \method improves accuracy for temporal tensors by exploiting temporal dependency and time-varying sparsity.
%
%
%


	\textbf{Applications.}
	Tensor decomposition have been used for various applications.
	\citet{KoldaBK05} analyze a hyperlink graph modeled as $3$-way tensor using CP decomposition.
	Tensor decomposition is also applied to tag recommendation~\Citep{RendleMNS09, RendleS10}.
	\citet{SunPLCLQ09} develop a content-based network analysis framework for finding higher-order clusters.
	\citet{LebedevGROL14} exploit CP decomposition to compress convolution filters of convolutional neural networks (CNNs).
Several works~\Citep{LeeOS18, PerrosPWVSTS17, PerrosPPVYdSS18} use tensor decomposition for analyzing Electronic Health Record (EHR) data.

\subsection{Tensor Factorization on Temporal Tensors}
	We explain tensor factorization methods dealing with temporal data.	
	\citet{dunlavy2011temporal} propose a tensor-based approach with an exponential smoothing technique for link prediction.
	\citet{matsubara2012fast} discover main topics in a complex temporal tensor and perform analysis for long periods of prediction.
	\citet{de2017tensorcast} present a non-negative coupled tensor factorization for forecasting future links in evolving networks.
	\citet{bahadori2014fast} propose an efficient low-rank tensor method to capture shared structures across multivariate spatial-temporal relationships.
	\citet{liu2019costco} propose a tensor completion method by exploiting the expressive power of convolutional neural networks to model non-linear interactions inside spatio-temporal tensors.
	These works focus on general multi faceted relationships rather than temporal characteristics.

	Several works~\Citep{xiong2010temporal,yu2016temporal,wu2019neural} model temporal patterns and trends in temporal tensors.
	\citet{xiong2010temporal} propose a Bayesian probabilistic tensor factorization method that learns global temporal patterns by adding an extra time factor to model evolving relations in a matrix.
	\citet{yu2016temporal} propose a matrix factorization method with an autoregressive temporal regularization to learn a temporal property.
	\citet{wu2019neural} propose a CP factorization method based on a long short-term memory network to model temporal interactions between latent factors of tensors.
	\citet{jing2018high} propose a tucker decomposition method
		to capture temporal correlations.
	However, these approaches are not designed for modeling temporal dependency from both past and future information,
	whereas \method obtains a time factor considering neighboring factors for both past and future time steps,
giving an accurate tensor factorization result.
	Moreover, they do not exploit the temporal sparsity, a common characteristic of a temporal tensor, while \method actively exploits the temporal sparsity.

\section{Conclusion}
\label{sec:conclusion}

We propose \method (\fullmethod), a novel tensor factorization method for temporal tensors
to accurately predict missing entries.
To capture temporal dependency and sparsity in real world temporal tensors,
we design a smoothing regularization on time factor,
and adjust the amount of the regularization according to the sparsity of time slices.
Moreover, we accurately and efficiently optimize \method with a carefully designed optimization strategy.
Extensive experimental results show that \method achieves up to $7.01\times$ lower RMSE and $5.50\times$ lower MAE compared to the second best methods.
Future works include extending \method for an online or a distributed setting.


\bibliographystyle{spbasic}
\bibliography{bib/dawon}

\begin{thebibliography}{45}
\providecommand{\natexlab}[1]{#1}
\providecommand{\url}[1]{{#1}}
\providecommand{\urlprefix}{URL }
\expandafter\ifx\csname urlstyle\endcsname\relax
  \providecommand{\doi}[1]{DOI~\discretionary{}{}{}#1}\else
  \providecommand{\doi}{DOI~\discretionary{}{}{}\begingroup
  \urlstyle{rm}\Url}\fi
\providecommand{\eprint}[2][]{\url{#2}}

\bibitem[{Acar et~al.(2011)Acar, Dunlavy, Kolda, and
  M{\o}rup}]{acar2011scalable}
Acar E, Dunlavy DM, Kolda TG, M{\o}rup M (2011) Scalable tensor factorizations
  for incomplete data. Chemometrics and Intelligent Laboratory Systems
  106(1):41--56

\bibitem[{Ahn et~al.(2020)Ahn, Son, and Kang}]{conf/cikm/AhnSK20}
Ahn D, Son S, Kang U (2020) Gtensor: Fast and accurate tensor analysis system
  using gpus. In: Proceedings of the 29th {ACM} International Conference on
  Information and Knowledge Management, {CIKM} 2020, October 19-23, 2020

\bibitem[{de~Araujo et~al.(2017)de~Araujo, Ribeiro, and
  Faloutsos}]{de2017tensorcast}
de~Araujo MR, Ribeiro PMP, Faloutsos C (2017) Tensorcast: Forecasting with
  context using coupled tensors (best paper award). In: 2017 IEEE International
  Conference on Data Mining (ICDM), IEEE, pp 71--80

\bibitem[{Bahadori et~al.(2014)Bahadori, Yu, and Liu}]{bahadori2014fast}
Bahadori MT, Yu QR, Liu Y (2014) Fast multivariate spatio-temporal analysis via
  low rank tensor learning. In: Advances in neural information processing
  systems, pp 3491--3499

\bibitem[{Battaglino et~al.(2018)Battaglino, Ballard, and
  Kolda}]{BattaglinoBK18}
Battaglino C, Ballard G, Kolda TG (2018) A practical randomized {CP} tensor
  decomposition. {SIAM} J Matrix Analysis Applications 39(2):876--901

\bibitem[{Beutel et~al.(2014)Beutel, Talukdar, Kumar, Faloutsos, Papalexakis,
  and Xing}]{BeutelTKFPX14}
Beutel A, Talukdar PP, Kumar A, Faloutsos C, Papalexakis EE, Xing EP (2014)
  Flexifact: Scalable flexible factorization of coupled tensors on hadoop. In:
  Proceedings of the 2014 {SIAM} International Conference on Data Mining,
  Philadelphia, Pennsylvania, USA, April 24-26, 2014, {SIAM}, pp 109--117

\bibitem[{Choi et~al.(2019)Choi, Jang, and Kang}]{10.1371/journal.pone.0217316}
Choi D, Jang JG, Kang U (2019) S3cmtf: Fast, accurate, and scalable method for
  incomplete coupled matrix-tensor factorization. PLOS ONE 14(6):1--20

\bibitem[{Choi and Vishwanathan(2014)}]{ChoiV14}
Choi JH, Vishwanathan S (2014) Dfacto: Distributed factorization of tensors.
  In: Ghahramani Z, Welling M, Cortes C, Lawrence ND, Weinberger KQ (eds)
  Advances in Neural Information Processing Systems 27: Annual Conference on
  Neural Information Processing Systems 2014, December 8-13 2014, Montreal,
  Quebec, Canada, pp 1296--1304

\bibitem[{Dunlavy et~al.(2011)Dunlavy, Kolda, and Acar}]{dunlavy2011temporal}
Dunlavy DM, Kolda TG, Acar E (2011) Temporal link prediction using matrix and
  tensor factorizations. ACM Transactions on Knowledge Discovery from Data
  (TKDD) 5(2):10

\bibitem[{Harshman et~al.(1970)}]{harshman1970foundations}
Harshman RA, et~al. (1970) Foundations of the parafac procedure: Models and
  conditions for an" explanatory" multimodal factor analysis

\bibitem[{Jeon et~al.(2016{\natexlab{a}})Jeon, Jeon, Sael, and
  Kang}]{conf/icde/JeonJSK16}
Jeon B, Jeon I, Sael L, Kang U (2016{\natexlab{a}}) Scout: Scalable coupled
  matrix-tensor factorization - algorithm and discoveries. In: 32nd {IEEE}
  International Conference on Data Engineering, {ICDE} 2016, Helsinki, Finland,
  May 16-20, 2016, pp 811--822

\bibitem[{Jeon et~al.(2015)Jeon, Papalexakis, Kang, and
  Faloutsos}]{jeon2015haten2}
Jeon I, Papalexakis EE, Kang U, Faloutsos C (2015) Haten2: Billion-scale tensor
  decompositions. In: 2015 IEEE 31st International Conference on Data
  Engineering, IEEE, pp 1047--1058

\bibitem[{Jeon et~al.(2016{\natexlab{b}})Jeon, Papalexakis, Faloutsos, Sael,
  and Kang}]{journals/vldb/JeonPFSK16}
Jeon I, Papalexakis EE, Faloutsos C, Sael L, Kang U (2016{\natexlab{b}}) Mining
  billion-scale tensors: algorithms and discoveries. {VLDB} J 25(4):519--544

\bibitem[{Jing et~al.(2018)Jing, Su, Jin, and Zhang}]{jing2018high}
Jing P, Su Y, Jin X, Zhang C (2018) High-order temporal correlation model
  learning for time-series prediction. IEEE transactions on cybernetics
  49(6):2385--2397

\bibitem[{Kang et~al.(2012)Kang, Papalexakis, Harpale, and
  Faloutsos}]{KangPHF12}
Kang U, Papalexakis EE, Harpale A, Faloutsos C (2012) Gigatensor: scaling
  tensor analysis up by 100 times - algorithms and discoveries. In: KDD, pp
  316--324

\bibitem[{Kiers(2000)}]{kiers2000towards}
Kiers HA (2000) Towards a standardized notation and terminology in multiway
  analysis. Journal of Chemometrics: A Journal of the Chemometrics Society
  14(3):105--122

\bibitem[{Kingma and Ba(2014)}]{kingma2014adam}
Kingma DP, Ba J (2014) Adam: A method for stochastic optimization. arXiv
  preprint arXiv:14126980

\bibitem[{Kolda and Sun(2008)}]{kolda2008scalable}
Kolda TG, Sun J (2008) Scalable tensor decompositions for multi-aspect data
  mining. In: 2008 Eighth IEEE international conference on data mining, IEEE,
  pp 363--372

\bibitem[{Kolda et~al.(2005)Kolda, Bader, and Kenny}]{KoldaBK05}
Kolda TG, Bader BW, Kenny JP (2005) Higher-order web link analysis using
  multilinear algebra. In: Proceedings of the 5th {IEEE} International
  Conference on Data Mining {(ICDM} 2005), 27-30 November 2005, Houston, Texas,
  {USA}, {IEEE} Computer Society, pp 242--249

\bibitem[{Lebedev et~al.(2015)Lebedev, Ganin, Rakhuba, Oseledets, and
  Lempitsky}]{LebedevGROL14}
Lebedev V, Ganin Y, Rakhuba M, Oseledets IV, Lempitsky VS (2015) Speeding-up
  convolutional neural networks using fine-tuned cp-decomposition. In: Bengio
  Y, LeCun Y (eds) 3rd International Conference on Learning Representations,
  {ICLR} 2015, San Diego, CA, USA, May 7-9, 2015, Conference Track Proceedings

\bibitem[{Lee et~al.(2018)Lee, Oh, and Sael}]{LeeOS18}
Lee J, Oh S, Sael L (2018) {GIFT:} guided and interpretable factorization for
  tensors with an application to large-scale multi-platform cancer analysis.
  Bioinform 34(24):4151--4158

\bibitem[{Liu et~al.(2019)Liu, Li, Tsang, and Liu}]{liu2019costco}
Liu H, Li Y, Tsang M, Liu Y (2019) Costco: A neural tensor completion model for
  sparse tensors. Training 10(4):10--3

\bibitem[{Maruhashi et~al.(2011)Maruhashi, Guo, and
  Faloutsos}]{maruhashi2011multiaspectforensics}
Maruhashi K, Guo F, Faloutsos C (2011) Multiaspectforensics: Pattern mining on
  large-scale heterogeneous networks with tensor analysis. In: 2011
  International Conference on Advances in Social Networks Analysis and Mining,
  IEEE, pp 203--210

\bibitem[{Matsubara et~al.(2012)Matsubara, Sakurai, Faloutsos, Iwata, and
  Yoshikawa}]{matsubara2012fast}
Matsubara Y, Sakurai Y, Faloutsos C, Iwata T, Yoshikawa M (2012) Fast mining
  and forecasting of complex time-stamped events. In: Proceedings of the 18th
  ACM SIGKDD international conference on Knowledge discovery and data mining,
  ACM, pp 271--279

\bibitem[{Oh et~al.(2018)Oh, Park, Sael, and Kang}]{OhPSK18}
Oh S, Park N, Sael L, Kang U (2018) Scalable tucker factorization for sparse
  tensors - algorithms and discoveries. In: 34th {IEEE} International
  Conference on Data Engineering, {ICDE} 2018, Paris, France, April 16-19, 2018

\bibitem[{Oh et~al.(2019)Oh, Park, Jang, Sael, and
  Kang}]{journals/tpds/OhPJSK19}
Oh S, Park N, Jang J, Sael L, Kang U (2019) High-performance tucker
  factorization on heterogeneous platforms. {IEEE} Trans Parallel Distrib Syst
  30(10):2237--2248

\bibitem[{Papalexakis et~al.(2012)Papalexakis, Faloutsos, and
  Sidiropoulos}]{PapalexakisFS12}
Papalexakis EE, Faloutsos C, Sidiropoulos ND (2012) Parcube: Sparse
  parallelizable tensor decompositions. In: ECML-PKDD, Springer, Lecture Notes
  in Computer Science, vol 7523, pp 521--536

\bibitem[{Park et~al.(2016)Park, Jeon, Lee, and Kang}]{conf/cikm/ParkJLK16}
Park N, Jeon B, Lee J, Kang U (2016) Bigtensor: Mining billion-scale tensor
  made easy. In: Proceedings of the 25th {ACM} International Conference on
  Information and Knowledge Management, {CIKM} 2016, Indianapolis, IN, USA,
  October 24-28, 2016, pp 2457--2460

\bibitem[{Park et~al.(2017)Park, Oh, and Kang}]{ParkOK17}
Park N, Oh S, Kang U (2017) Fast and scalable distributed boolean tensor
  factorization. In: 33rd {IEEE} International Conference on Data Engineering,
  {ICDE} 2017, San Diego, CA, USA, April 19-22, 2017, pp 1071--1082

\bibitem[{Park et~al.(2019)Park, Oh, and Kang}]{vldbj/Park2019}
Park N, Oh S, Kang U (2019) Fast and scalable method for distributed boolean
  tensor factorization. The VLDB Journal

\bibitem[{Perros et~al.(2017)Perros, Papalexakis, Wang, Vuduc, Searles,
  Thompson, and Sun}]{PerrosPWVSTS17}
Perros I, Papalexakis EE, Wang F, Vuduc RW, Searles E, Thompson M, Sun J (2017)
  Spartan: Scalable {PARAFAC2} for large {\&} sparse data. In: Proceedings of
  the 23rd {ACM} {SIGKDD} International Conference on Knowledge Discovery and
  Data Mining, Halifax, NS, Canada, August 13 - 17, 2017, {ACM}, pp 375--384

\bibitem[{Perros et~al.(2018)Perros, Papalexakis, Park, Vuduc, Yan, deFilippi,
  Stewart, and Sun}]{PerrosPPVYdSS18}
Perros I, Papalexakis EE, Park H, Vuduc RW, Yan X, deFilippi C, Stewart WF, Sun
  J (2018) Sustain: Scalable unsupervised scoring for tensors and its
  application to phenotyping. In: Guo Y, Farooq F (eds) Proceedings of the 24th
  {ACM} {SIGKDD} International Conference on Knowledge Discovery {\&} Data
  Mining, {KDD} 2018, London, UK, August 19-23, 2018, {ACM}, pp 2080--2089

\bibitem[{Rendle and Schmidt{-}Thieme(2010)}]{RendleS10}
Rendle S, Schmidt{-}Thieme L (2010) Pairwise interaction tensor factorization
  for personalized tag recommendation. In: WSDM, pp 81--90

\bibitem[{Rendle et~al.(2009)Rendle, Marinho, Nanopoulos, and
  Schmidt{-}Thieme}]{RendleMNS09}
Rendle S, Marinho LB, Nanopoulos A, Schmidt{-}Thieme L (2009) Learning optimal
  ranking with tensor factorization for tag recommendation. In: SIGKDD, pp
  727--736

\bibitem[{Shin and Kang(2014)}]{conf/icdm/ShinK14}
Shin K, Kang U (2014) Distributed methods for high-dimensional and large-scale
  tensor factorization. In: 2014 {IEEE} International Conference on Data
  Mining, {ICDM} 2014, Shenzhen, China, December 14-17, 2014, pp 989--994

\bibitem[{Shin et~al.(2016)Shin, Sael, and Kang}]{shin2016fully}
Shin K, Sael L, Kang U (2016) Fully scalable methods for distributed tensor
  factorization. IEEE Transactions on Knowledge and Data Engineering
  29(1):100--113

\bibitem[{Smith and Karypis(2017)}]{SmithK17}
Smith S, Karypis G (2017) Accelerating the tucker decomposition with compressed
  sparse tensors. In: Rivera FF, Pena TF, Cabaleiro JC (eds) Euro-Par 2017:
  Parallel Processing - 23rd International Conference on Parallel and
  Distributed Computing, Santiago de Compostela, Spain, August 28 - September
  1, 2017, Proceedings, Springer, Lecture Notes in Computer Science, vol 10417,
  pp 653--668

\bibitem[{Sun et~al.(2006)Sun, Papadimitriou, and Philip}]{sun2006window}
Sun J, Papadimitriou S, Philip SY (2006) Window-based tensor analysis on
  high-dimensional and multi-aspect streams. In: Sixth International Conference
  on Data Mining (ICDM'06), IEEE, pp 1076--1080

\bibitem[{Sun et~al.(2009)Sun, Papadimitriou, Lin, Cao, Liu, and
  Qian}]{SunPLCLQ09}
Sun J, Papadimitriou S, Lin C, Cao N, Liu S, Qian W (2009) Multivis:
  Content-based social network exploration through multi-way visual analysis.
  In: Proceedings of the {SIAM} International Conference on Data Mining, {SDM}
  2009, April 30 - May 2, 2009, Sparks, Nevada, {USA}, {SIAM}, pp 1064--1075

\bibitem[{Sun et~al.(2015)Sun, Gao, Hong, Mishra, and
  Yin}]{sun2015heterogeneous}
Sun Y, Gao J, Hong X, Mishra B, Yin B (2015) Heterogeneous tensor decomposition
  for clustering via manifold optimization. IEEE transactions on pattern
  analysis and machine intelligence 38(3):476--489

\bibitem[{Symeonidis(2016)}]{symeonidis2016matrix}
Symeonidis P (2016) Matrix and tensor decomposition in recommender systems. In:
  Proceedings of the 10th ACM Conference on Recommender Systems, pp 429--430

\bibitem[{Wu et~al.(2019)Wu, Shi, Dong, Huang, and Chawla}]{wu2019neural}
Wu X, Shi B, Dong Y, Huang C, Chawla NV (2019) Neural tensor factorization for
  temporal interaction learning. In: Proceedings of the Twelfth ACM
  International Conference on Web Search and Data Mining, pp 537--545

\bibitem[{Xiong et~al.(2010)Xiong, Chen, Huang, Schneider, and
  Carbonell}]{xiong2010temporal}
Xiong L, Chen X, Huang TK, Schneider J, Carbonell JG (2010) Temporal
  collaborative filtering with bayesian probabilistic tensor factorization. In:
  Proceedings of the 2010 SIAM International Conference on Data Mining, SIAM,
  pp 211--222

\bibitem[{Yu et~al.(2016)Yu, Rao, and Dhillon}]{yu2016temporal}
Yu HF, Rao N, Dhillon IS (2016) Temporal regularized matrix factorization for
  high-dimensional time series prediction. In: Advances in neural information
  processing systems, pp 847--855

\bibitem[{Zhang et~al.(2017)Zhang, Guo, Dong, He, Xu, and
  Chen}]{zhang2017cautionary}
Zhang S, Guo B, Dong A, He J, Xu Z, Chen SX (2017) Cautionary tales on
  air-quality improvement in beijing. Proceedings of the Royal Society A:
  Mathematical, Physical and Engineering Sciences 473(2205):20170457

\end{thebibliography}

\appendix
\section{Appendix}

\subsection{Row-wise Update Rule} \label{subsec:rowwiseupdate}
We use a row-wise update rule to efficiently update non-time factor matrices.
This update rule has an advantage of considering only nonzero values and allows easy parallelization. 
Following the notations and equations introduced by~\cite{shin2016fully}, 
the update rule for $i_n$th row of the $n$th factor matrix $\A{n} (n \neq t)$ is given as follows:
	\begin{align} \label{eq:ls}
		\begin{split}
		[a_{i_n 1}^{(n)}, \cdots, a_{i_n K}^{(n)}] \leftarrow
		\argmin{[a^{(n)}_{i_{n}1}, ..., a^{(n)}_{i_{n}K}]}{L(\mat{A}^{(1)},...,\mat{A}^{(N)})} \\
		= \vect{c}_{i_n:}^{(n)} \times [\mat{B}_{i_n}^{(n)}+\lambda_r \mathbf{I}_{K}]^{-1}\text{\space\space}
		\end{split}
		\end{align}
	where $\mat{B}_{i_n}^{(n)}$ is a  ${K \times K}$ matrix whose entries are
	\begin{equation} \label{eq:rowB}
		{(\mat{B}_{i_n}^{(n)})}_{k_1 k_2} =
	 	\sum_{\forall\alpha\in\Omega_{i_n}^{(n)}}
		\prod_{l \neq n} a^{(l)}_{i_l k_1}\prod_{l \neq n} a^{(l)}_{i_l k_2}, \forall k_1, k_2
	\end{equation}
, $\vect{c}_{i_n}^{(n)}$ is a length ${K}$ vector whose entries are
	\begin{equation} \label{eq:rowC}
		\sum_{\forall\alpha\in\Omega_{i_n}^{(n)}}x_{\alpha}\prod_{l \neq n} a^{(l)}_{i_l k}, \forall k
	\end{equation}
	and $\Omega_{i_n}^{(n)}$ denotes the subset of $\Omega$ whose $n$th mode's index is $i_n$.

\end{document}